\documentclass{article}

\PassOptionsToPackage{numbers, compress}{natbib}

\usepackage[final]{neurips_2025}

\usepackage{amsthm}
\usepackage{amsmath, amssymb}
\usepackage{algorithm}
\usepackage{algpseudocode}
\usepackage{xcolor}  
\usepackage{mathtools}
\usepackage{todonotes}
\usepackage{graphicx}    
\usepackage{makecell}

\newcommand{\vb}{{\boldsymbol\beta}}

\newcommand{\vt}{{\boldsymbol\theta}}
\newcommand{\vp}{{\boldsymbol\psi}}
\newcommand{\vm}{\boldsymbol{\mu}}
\newcommand{\vx}{\mathbf{x}}
\newcommand{\vy}{\mathbf{y}}
\newcommand{\vz}{\mathbf{z}}

\newcommand{\vw}{\mathbf{w}}

\newcommand{\vW}{\mathbf{W}}
\newcommand{\vM}{\mathbf{M}}
\newcommand{\vX}{\mathbf{X}}
\newcommand{\vY}{\mathbf{Y}}
\newcommand{\vZ}{\mathbf{Z}}

\newcommand{\mG}{\mathcal{G}}
\newcommand{\mF}{\mathcal{F}}
\newcommand{\mH}{\mathcal{H}}
\newcommand{\mX}{\mathcal{X}}
\newcommand{\mR}{\mathcal{R}}

\newcommand{\mZ}{\mathcal{Z}}

\usepackage[utf8]{inputenc} 
\usepackage[T1]{fontenc}    
\usepackage{hyperref}       
\usepackage{cleveref}
\Crefname{equation}{Eq.}{Eqs.}
\usepackage{url}            
\usepackage{booktabs}       
\usepackage{amsfonts}       
\usepackage{amsmath}
\usepackage{nicefrac}       
\usepackage{microtype}      
\usepackage{wrapfig}
\usepackage{subcaption}
\usepackage{xcolor}         
\usepackage{colortbl} 
\usepackage{multirow}
\PassOptionsToPackage{table}{xcolor}
\usepackage{tcolorbox} 
\tcbuselibrary{skins, breakable}
\colorlet{lightblue}{blue!10!white}

\setcitestyle{square}

\newcommand*{\algname}{H-SPLID}

\newcommand{\kclass}{k}
\newcommand{\inputspace}{\mathcal{X}}
\newcommand{\labelspace}{\mathcal{Y}}
\newcommand{\kxspace}{\mathcal{F}}
\newcommand{\kzspace}{\mathcal{G}}
\newcommand{\dataset}{\mathcal{D}}
\newcommand{\inputvec}{\vx}

\newcommand{\labelvec}{\vy}
\newcommand{\numdata}{n}
\newcommand{\encoder}{f_\vp}

\newcommand{\latentrep}{\vz}
\newcommand{\latentspace}{\mathcal{R}}
\newcommand{\encparam}{\vp}
\newcommand{\encparamdim}{p}
\newcommand{\logits}{g_{\vW}}
\newcommand{\logitparam}{\vW}
\newcommand{\logitparamdim}{k \times m}

\newcommand{\loss}{\mathcal{L}}

\definecolor{deemph}{gray}{0.6}

\newtheorem{theorem}{Theorem}[section]
\newtheorem{lemma}[theorem]{Lemma}
\newtheorem{corollary}[theorem]{Corollary}
\newtheorem{proposition}[theorem]{Proposition}

\theoremstyle{definition}

\newtheorem{assumption}[theorem]{Assumption}

\title{H-SPLID: HSIC-based Saliency Preserving  Latent Information Decomposition}

\author{
Lukas Miklautz\textsuperscript{1, $\dagger$,}\thanks{Equal contribution. \textsuperscript{$\ddagger$}Shared supervision. \textsuperscript{$\dagger$}Main work done during a research stay at Northeastern University.}\quad
{Chengzhi Shi}\textsuperscript{2,}\footnotemark[1]\quad
Andrii Shkabrii\textsuperscript{3,4,}\footnotemark[1]\quad
Theodoros Thirimachos Davarakis\textsuperscript{2}\\
\textbf{Prudence Lam}\textsuperscript{2}\quad
\textbf{Claudia Plant}\textsuperscript{3, 5, $\ddagger$}\quad
\textbf{Jennifer Dy}\textsuperscript{2, $\ddagger$}\quad
\textbf{Stratis Ioannidis}\textsuperscript{2, $\ddagger$}\\
\textsuperscript{1}Department of Machine Learning and Systems Biology, Max Planck Institute\\ of Biochemistry, Martinsried, Germany 
\textsuperscript{2}Northeastern University, Boston, MA, USA \\
\textsuperscript{3}Faculty of Computer Science, University of Vienna, Vienna, Austria \\
\textsuperscript{4}Doctoral School Computer Science, University of Vienna, Vienna, Austria \\
\textsuperscript{5}Research Network Data Science, University of Vienna, Vienna, Austria
}

\begin{document}

\maketitle

\begin{abstract}
We introduce \algname{}, a novel algorithm for learning salient feature representations through the explicit decomposition of salient and non-salient features into separate spaces. We show that \algname{} promotes learning low-dimensional, task-relevant features. We prove that the expected prediction deviation under input perturbations is upper-bounded by the dimension of the salient subspace and the Hilbert-Schmidt Independence Criterion (HSIC) between inputs and representations. This  establishes a link between robustness and latent representation compression in terms of the dimensionality and information preserved.
Empirical evaluations on image classification tasks show that models trained with \algname{} primarily rely on salient input components, as indicated by reduced sensitivity to perturbations affecting non-salient features, such as image backgrounds.
\end{abstract}

\section{Introduction}
\label{sec:introduction}

\begin{figure}[t]
    \centering
\includegraphics[width=0.85\textwidth]{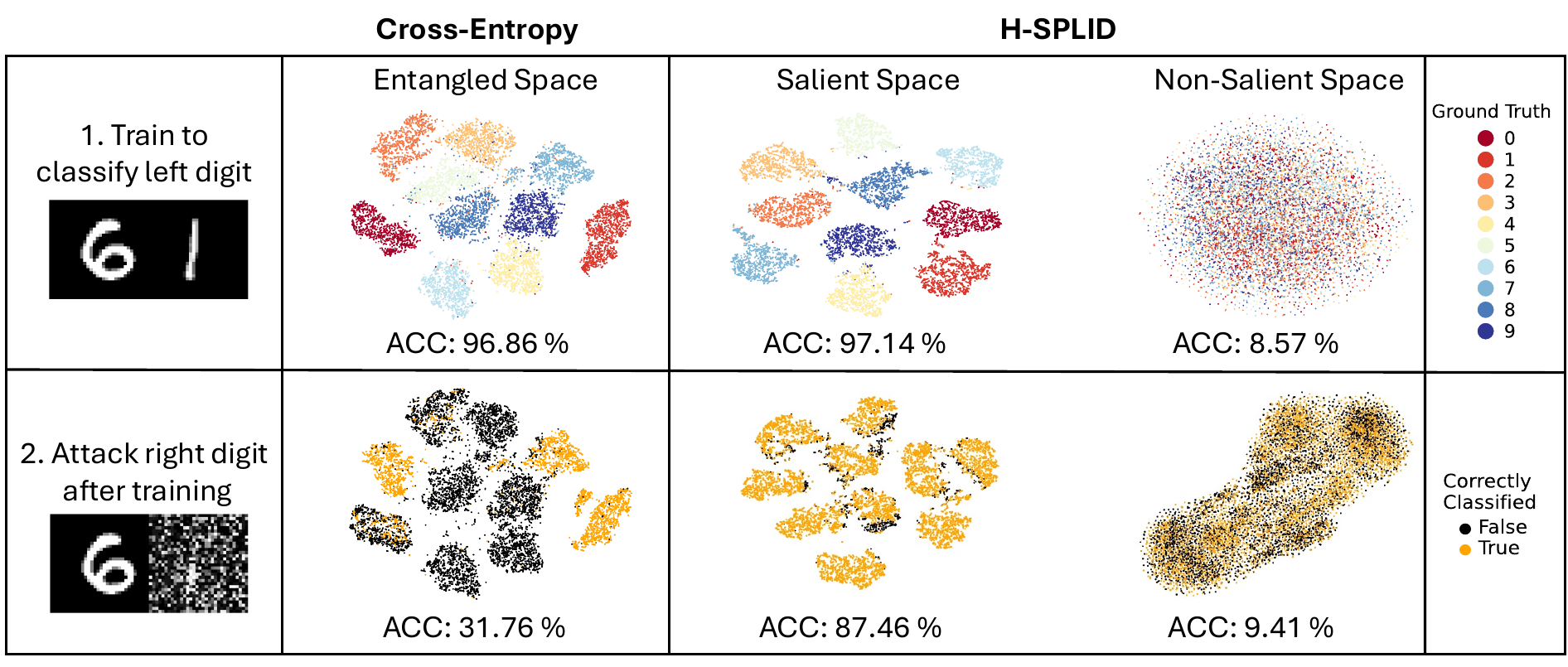}
\caption{
\textbf{H-SPLID learns to ignore irrelevant input by decomposing the latent space into salient and non-salient components.}
\textbf{Left:} A simple diagnostic test for saliency, where the model is trained to classify the left digit (only labels for the left are provided) and it should ignore the right.
\textbf{Middle:} A model trained with \emph{cross-entropy} loss achieves high test accuracy (96.86\%) but produces entangled representations, making it sensitive to perturbations on the right digit (accuracy drops to 31.76\% under high-magnitude PGD attack).
\textbf{Right:} \algname{} separates the latent space into a salient subspace, which captures class-discriminative structure (ACC 97.14\%), and a non-salient subspace, which contains no class-relevant information (ACC 8.57\%). This separation enables robustness to perturbations on irrelevant input (ACC 87.46\%), showing proper learning of salient features. Embeddings are visualized using t-SNE \citep{tsne}.
    }
    \label{fig:motivation}
\end{figure}

The acquisition of salient, task-relevant features from high-dimensional inputs constitutes a fundamental challenge in representation learning. Such features offer multiple advantages, including reduced dimensionality \citep{Abid2018}, enhanced generalization and transferability \citep{ros, pmlr-v37-long15}, and improved robustness \citep{ismail2021improving, etmann}. Nevertheless, learning true salient features remains challenging, as many neural networks operate within a single, entangled latent space that mixes task-relevant signals with redundant information \citep{Bengio2012RepresentationLA, montero2022lost}. 
We illustrate this sensitivity using a simple diagnostic test in Figure~\ref{fig:motivation}: a classifier trained to predict the left digit in an image of double digits should ignore perturbations to the right digit, which is irrelevant to the label. However, in practice, we observe that neural networks with high test classification accuracy on the left digits exhibit a significant performance drop when subjected to a high-magnitude adversarial PGD \citep{pgd} attack ($\epsilon=1.0$) on the right digits, revealing their dependence on irrelevant, non-salient features.  This corroborates several empirical and theoretical studies \citep{vib, melamed2023adversarial, haldar2024effect} showing that redundant  dimensions enhance vulnerability to attacks.
Driven by these findings, we introduce \algname{} (HSIC-based Saliency-Preserving Latent Information Decomposition), a new method that learns salient features by explicitly decomposing the latent space into two subspaces coupled with information compression regularization: a low-dimensional \emph{salient space}, which contains features essential for classification, and a \emph{non-salient space}, which captures the remaining input variability. 
Training the same neural network as above with \algname{} alleviates the dependence on irrelevant features to a large degree, as shown on the right side of Figure \ref{fig:motivation}. Crucially, \algname{} is significantly less sensitive to right digit perturbations, without any prior knowledge of the redundant region or adversarial training. 


We extend this analysis in our experiments by applying adversarial attacks to the background of COCO \citep{coco} images, applying image corruptions to medical images of skin lesions \citep{isic2018skin} and further show that \algname{} improves the transfer accuracy of ResNet-based~\cite{resnet} classifiers trained on ImageNet \citep{imagenet} under real-world perturbations \citep{imagenet_9, counteranimal}. In addition to our empirical evidence, we theoretically prove that the expected change in predictions under input perturbations is bounded by the dimension of the learned salient subspace and the Hilbert-Schmidt Independence Criterion (HSIC) \citep{gretton2005measuring} between inputs and salient representations. This establishes a formal link between robustness and the salient representation. Our main contributions are:
\begin{itemize}
    \item We propose \algname{}, a novel algorithm that promotes the learning of salient features by decomposing the network's latent space into salient and non-salient subspaces.
    \item  We prove that the two key design components of \algname{}, namely, dimensionality reduction of the salient subspace, together with the HSIC between inputs and salient latent representations, upper bounds the expected change in predictions under input perturbations. Moreover, we show that the above HSIC and reduced salient space dimensionality bounds the volume of the input domain that is vulnerable to perturbations.
    \item We empirically demonstrate that \algname{} learns salient features by leveraging attacks and other perturbations against non‐salient regions of an image, such as its background.
\end{itemize}

\section{Related Work}
\label{sec:related_work}




\paragraph{Salient Feature Learning.}

Saliency methods in interpretability aim to identify input features that influence a model's prediction the most. Traditional post hoc approaches include gradient-based methods \citep{SimonyanVZ13, intgrad}, Class Activation Maps (CAMs) \citep{gradcam, wang2020score}, and perturbation-based methods \citep{shap, lime}. Unlike post hoc interpretability methods, however, \algname{} aims to learn latent salient features for a given task, such as image classification.
%
Existing works on salient feature learning include  saliency-guided training for interpretability \citep{ismail2021improving} and saliency-based data augmentation \citep{chen2023saliencyguidedcontrastivelearning,aydemir2024dataaugmentationlatentdiffusion, uddin2021saliencymix} methods that can complement our approach. However, \algname{} does not use saliency maps as an auxiliary signal to improve training \citep{chen2023saliencyguidedcontrastivelearning, uddin2021saliencymix}, or pretrained models \citep{aydemir2024dataaugmentationlatentdiffusion} to generate them. Moreover, the division of the latent space into ``salient'' and ``non-salient'' spaces, as in \algname{}, is comparatively unexplored in literature. Contrastive Analysis (CA) methods \citep{Abid2018,Abid2019ContrastiveVA, weinbergerBL22} leverage this concept by learning explicit ``common'' and ``salient'' latent spaces with separate encoders via external supervision. While they share the idea of learning separate spaces with \algname{}, they rely on a dedicated target dataset containing the salient class, and a background dataset with samples exhibiting non-salient features. In contrast, our method does not rely on external data, and learns an initial unified latent space, before partitioning it into salient and non-salient dimensions.

\paragraph{Feature Decomposition and Selection.}
Feature selection methods aim to identify a subset of input features that is most predictive of the target variable \cite{feature_select, GuyonE03}, with popular approaches including \emph{\(L_1\)} regularization \citep{lasso} and \emph{Group-Lasso} regularization \citep{group_lasso}. Similar to these methods, \algname{} transforms and selects features during training, but diverges in its approach through its decomposition of the latent space. For the task of clustering, \citet{ijcai2021p389} recently introduced the idea of latent space partitioning, 
whereas \algname{} embeds this split directly into a classifier's training loop, using labels to shape the salient vs.~non-salient space. Moreover, \algname{} incorporates the HSIC penalty \citep{hsic_bottleneck} to regularize the statistical dependence between the inputs and salient features from each subspace, ensuring that the salient subspace retains only task-relevant information while the non-salient subspace absorbs redundant variability, thereby reducing feature dimensions.

\paragraph{Adversarial Robustness and Saliency.}
We use adversarial attacks to evaluate the quality of the learned salient features. Several studies have begun exploring the interplay between saliency and robustness \citep{etmann, tsipras2018robustness, guesmi2024exploringinterplayinterpretabilityrobustness, li_ijcai}. Among these methods, many require adversarial training \citep{pgd}, which is not only computationally demanding, but also tailored to specific attacks. An alternative line of research seeks to enhance robustness without adversarial training. Multiple works \citep{haldar2024effect,vib,hbar,ceb} have attributed adversarial vulnerability to the network's reliance on high-dimensional, task-irrelevant features. For instance, \citet{vib} hypothesize that neural networks falsely rely on task-irrelevant features from the training data, negatively impacting robust generalization. \citet{melamed2023adversarial} show, under a simplified two-layer model, that when data is confined to a low-dimensional manifold, there exists an off-manifold space in which weights remain mostly unchanged and can be exploited by adversarial perturbations. 
\citet{haldar2024effect} demonstrate that when there are redundant latent dimensions, off-manifold attacks can lead to decision boundaries that rely on task-irrelevant feature dimensions. \citet{ceb} introduced an information bottleneck \citep{tishby} to compress input information and preserve task-relevant features  without adversarial training. \citet{hbar} extends this framework to the HSIC bottleneck \citep{hsic_bottleneck}, upon which \algname{} is built. Importantly, \algname{} departs from the HSIC bottleneck penalty, and introduces separate terms for salient and non-salient features. Our method also improves upon the guarantees of \citet{hbar}, tightening their bounds to account for the impact of the dimensionality reduction induced by \algname{}.

\section{Methodology}\label{sec:method}

\subsection{Problem Setup}
\label{sec:problem_setup}
We consider  $\kclass$-class classification over  dataset  $\dataset = \{(\inputvec_i, \labelvec_i)\}_{i=1}^{\numdata} \subseteq \inputspace \times \labelspace$,  where $\inputspace \subseteq \mathbb{R}^d$ is a compact input space, $\labelspace \subseteq \mathbb{R}^k$ is the label space, and each input $\inputvec_i \in \inputspace$ is associated with the corresponding one-hot class label $\labelvec_i \in \labelspace$.  
The neural network  $h_\vt : \mathbb{R}^d \to \mathbb{R}^k$ consists of an encoder followed by a linear layer. The encoder, denoted by $\encoder: \inputspace \to \latentspace$ with parameters $\encparam \in \mathbb{R}^{\encparamdim}$, maps an input $\inputvec$ to a latent representation $\latentrep \in \latentspace \subseteq \mathbb{R}^m$, i.e., \(\latentrep = \encoder(\inputvec)\). The linear output layer, $\logits: \latentspace \to \mathbb{R}^\kclass$, computes the $\kclass$ logits using parameters $\logitparam \in \mathbb{R}^{\logitparamdim}$, i.e., \(\logits(\latentrep) = \logitparam \latentrep\). 
Thus, we can express the neural network $h_\vt= \logits \circ \encoder $ with parameters $\vt = \{\encparam, \logitparam\}$. Hence, the prediction of sample $i$ is $\hat{\vy}_i = \vW f_\vp(\vx_i)$. Lastly, parameters  $\vt = \{\encparam, \logitparam\}$ are trained by minimizing the cross-entropy loss with a softmax layer, i.e., $\loss_{ce}(\dataset; \vt) = \frac{1}{\numdata} \sum_{i=1}^{\numdata} \ell(\inputvec_i, \labelvec_i, \vt),$ where $\ell(\inputvec, \labelvec; \vt) = - \sum_{j=1}^k y_j\log(\sigma_j(\hat{\vy}))$ and  $\sigma_i(\vy)=e^{\vy_i}/\sum_je^{\vy_j}$ is the softmax function $\sigma:\mathbb{R}^k\to\mathbb{R}^k$.





\subsection{Saliency-Aware Latent Decomposition}
\begin{figure}[t]
    \centering
    \includegraphics[width=0.9\textwidth]{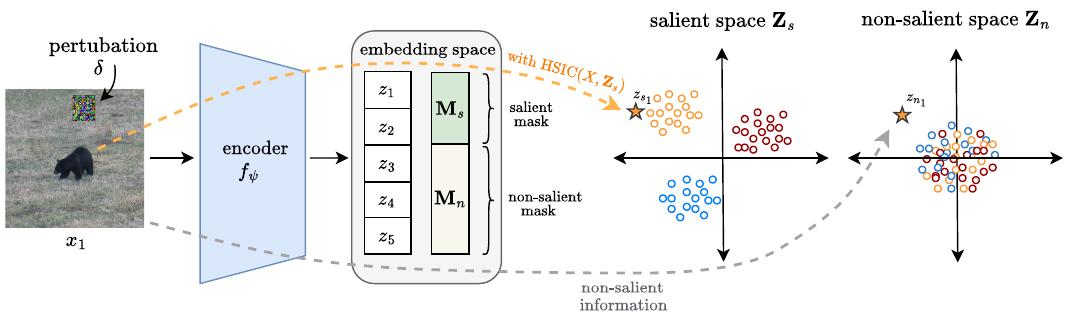}
    \caption{\textbf{Overview of  \algname{}.} The salient information for classifying the black bear is encoded in the salient space $\vz_s$, whereas the background information is encoded in the non-salient space $\vz_n$, allowing \algname{} to be more robust to perturbations $\delta$ of the background.
        }
    \label{fig:method_overview}
\end{figure}
We introduce a representation learning framework that separates latent features into \emph{salient} (i.e., task-relevant) and \emph{non-salient} (i.e., task-irrelevant) components. The model is trained with a structured objective that integrates classification, geometric regularization, and statistical independence constraints, producing representations that improve both predictive performance and robustness. An overview of our method is shown in Figure \ref{fig:method_overview}
.

Given the encoder $f_\vp$, we introduce a learnable diagonal mask matrix \(\vM_s =\operatorname{diag}\{\vb\} \in \{0,1\}^{m \times m}\), where $\vb\in \{0,1\}^m$ selects salient (task-relevant) features. The complementary non-salient mask is defined as \(\vM_n = \mathbf{I} - \vM_s\).
The latent representation \(\latentrep = f_\vp(\inputvec) \in \mR\) is then decomposed into the following salient and non-salient representations: \( \latentrep_s = \vM_s \latentrep =\vb \odot \latentrep\) and \( \latentrep_n = \vM_n\latentrep = (\mathbf{1}- \vb) \odot \latentrep\), with \(\mathbf{1}\) being a vector of $1$.


Define $(\vy_i)_j$ as the $j$-th element of the $\vy_i$ label. Then, classification is performed using \emph{only the salient component of the latent space} $\latentrep_s\in \mathcal{Z}$, and the corresponding cross-entropy loss is given by:
\begin{equation}\label{eq: ce loss}
\loss_{\text{ce}}(\dataset; \vt, \vM_s) = - \frac{1}{n} \sum_{i=1}^n (\vy_i)_j\log(\sigma(\hat{\vy}_i)), \quad \text{with} \quad \hat{\vy} = \logitparam^\top \vM_s f_\vp(\inputvec_i).
\end{equation}
\subsection{Regularizing and Preserving the Separated Features}

While the saliency masks enable latent space partitioning, the quality and stability of this separation depend on additional constraints that encourage discriminative utility. We achieve this via two regularization mechanisms: masked clustering losses and Hilbert-Schmidt Independence Criterion (HSIC) \cite{gretton2005measuring} penalties.

Let $\vX=[\vx_1, \dots, \vx_n]\in \mathbb{R}^{d \times n}$ be the matrix of input vectors and $\vZ\equiv f_\vp(\vX) \in \mathbb{R}^{m \times n}$ be the matrix of latent vectors for $n$ samples, with masked variants $\vz_s = \vM_s \vZ$ and $\vZ_n = \vM_n \vZ$. For each class $k$, let $C_k$ denote the set of indices with label $k$, and define $\vm_k$ as the class centroid and $\vm$ as the global centroid of latent vectors. We define the following masked norm-based losses  \citep{ijcai2021p389}:
\begin{equation}\label{eq: acdc loss}
\loss_s(\dataset;\vp, \vM_s) = \sum_{k=1}^K \sum_{i \in C_k} \|\vM_s(\vz_i - \vm_k)\|^2, \qquad
\loss_n(\dataset;\vp, \vM_n) = \sum_{i=1}^n \|\vM_n(\vz_i - \vm)\|^2.
\end{equation}

Loss $\loss_s$ encourages the clustering of class-specific representations in the salient subspace, strengthening its discriminative capacity. Each class has a simple uni-modal form which further removes redundant information. The loss $\loss_n$ aligns the other features globally across samples in the non-salient space and captures shared variation. Moreover, by capturing task-irrelevant variations—such as background information—non-salient features help isolate predictive factors in the salient space, enhancing robustness and disentanglement.
To further promote robust but accurate decompositions, we incorporate two additional HSIC terms:
\begin{align}\label{eq: hsic loss}
\rho_s \widehat{\operatorname{HSIC}}(\vX, \vZ_s) + \rho_n \widehat{\operatorname{HSIC}}(\vY, \vZ_n).
\end{align}
As HSIC is a measure of similarity (see \Cref{app: bg kn}), the term $\widehat{\operatorname{HSIC}}(\vX, \vZ_s)$ reduces the dependence between the input features and the salient subspace, thereby removing redundant information. Similarly, \(\widehat{\operatorname{HSIC}}(\vY, \vZ_n)\) reduces the dependence between label information in the non-salient subspace. 
We use  the unbiased empirical estimator \citep{gretton2005measuring} of HSIC:
\begin{align}\widehat{\operatorname{HSIC}}(\vX,\vZ_s)=\sum_{j=1}^k \sum_{i=1}^k (n-1)^{-2} \operatorname{tr}(K_x H K_z H),
\end{align}
where $K_x$ and $K_z $ have elements $K_{x_{(i,j)}}=k_x(\vx_i, \vx_j)$ and $K_{z_{(i,j)}}=k_z(\vM_s\vz_{i}, \vM_s\vz_{j})$, and $H=\mathbf{I}-\frac{1}{n} \mathbf{1} \mathbf{1}^\top$ is the centering matrix.


Putting everything together, we define \algname{}'s overall training objective as:
\begin{align}\label{eq: st loss}
\mathcal{L}(\mathcal{D}; \vt, \vM_s, \vM_n) = 
& \lambda_{ce}\mathcal{L}_{\text{ce}}(\mathcal{D}; \vt, \vM_s)
+ \lambda_s \mathcal{L}_s(\mathcal{D}; \vp, \vM_s)
+ \lambda_n \mathcal{L}_n(\mathcal{D}; \vp, \vM_n) \nonumber \\
& + \rho_s \operatorname{HSIC}(\vX, \vZ_s)
+ \rho_n \operatorname{HSIC}(\vY, \vZ_n),
\end{align}
where \( \vZ_s, \vZ_n \in \mathbb{R}^{m \times n} \) are the concatenated salient and non-salient latent representation, and \(\lambda_{ce}, \lambda_s, \lambda_n, \rho_s, \rho_n \geq 0 \) are scalar weights. 
Training amounts to solving the following constrained optimization problem:
\begin{subequations}
\label{eq:prob}
\begin{align}
\min_{\vt, \vM_s} \quad & \mathcal{L}(\mathcal{D}; \vt, \vM_s, \vM_n) \\
\text{subject to} \quad 
& \vM_n = I - \vM_s,\quad \vz_i = f_\vp(\vx_i), ~ \forall i \in \{1, \dots, n\}. \\
& \boldsymbol{\mu}_k = \frac{1}{|C_k|} \sum_{i \in C_k} \vz_i, \quad
\boldsymbol{\mu} = \frac{1}{n} \sum_{i=1}^n \vz_i. \label{eq: means}
\end{align}
\end{subequations}


\subsection{The \algname{} Algorithm}\label{sec:altopt}
We solve  Problem \eqref{eq:prob} using an alternating optimization procedure over the neural network parameters \( \vt \) and the diagonal mask matrix \( \vM_s \in \mathbb{R}^{m \times m} \) (see Algorithm~\ref{alg:alt-opt} in \Cref{app: psd code}). At each outer epoch \(t\), the procedure consists of two alternating steps:

\textbf{(a) Latent Representation Update (Fix \( \vM_s \), optimize \( \vt\))}:
Given a fixed mask \( \vM_s^{(t-1)} \), we update the encoder parameters \( \vt = \{ \vp, \vW\} \) by minimizing the loss \( \mathcal{L} \) as in \Cref{eq: st loss} using minibatch stochastic gradient descent with \( \mathcal{B} \subset \mathcal{D} \) for an epoch:
\[
\vt^{(t)}\leftarrow \vt^{t-1} - \eta \nabla_{\vt} \mathcal{L} (\mathcal{B}; \vt^{(t-1)}, \vM_s^{(t-1)}, \vM_n^{(t-1)})
\]
where the class means $\vm_k$ and global means $\vm$ are computed based on the minibatch via \Cref{eq: means}. 

\textbf{(b) Mask Update (Fix \( \vt \), optimize \( \vM_s \))}:
With updated latent representations \( \vz_i^{(t)} = f_{\vp^{(t)}}(\vx_i) \), we optimize the following optimization problem to learn the masks $\vM_s^{(t)}, \vM_n^{(t)}$:
\begin{subequations}\label{eq:maskprob}
\begin{align}
    \min_{\vb} \quad & \lambda_s \mathcal{L}_s(\vZ^{(t)}, \vM_s) + \lambda_n \mathcal{L}_n(\vZ^{(t)}, \vM_n)\\
    \text{subject to} \quad & \vM_s = \operatorname{diag}\{\vb\}, \quad \vM_n = \mathbf{I}-\vM_s. 
\end{align}
\end{subequations}
 \citet{ijcai2021p389} show that Prob.~\ref{eq:maskprob} has a closed-form solution:
\[
\vb_i^* = \frac{
\lambda_n \sum_{\vz \in \mathcal{D}} (\vz_i - \boldsymbol{\mu}_i)^2
}{
\lambda_s \sum_{k=1}^K \sum_{\vz \in C_k} (\vz_i - (\boldsymbol{\mu}_k)_i)^2 +
\lambda_n \sum_{\vz \in \mathcal{D}} (\vz_i - \boldsymbol{\mu}_i)^2
},
\quad \forall i \in \{1, \dots, m\},
\]
where the global mean \( \boldsymbol{\mu} \) and class means \( \boldsymbol{\mu}_k \) are computed from the full dataset via \Cref{eq: means}. We  use a moving average when updating the masks to improve convergence (See Algorithm~\ref{alg:alt-opt} line 11 in \Cref{app: psd code}). The above optimization yields a continuous mask $\vb \in [0,1]^m$. After convergence, we obtain a binary version by thresholding each entry at $0.5$. In practice, using the continuous mask directly gives similar results, as the learned values typically concentrate near $0$ or $1$.


\subsection{Theoretical Guarantees}
\label{subsec: theory}

In our experiments, we study whether \algname{} relies on salient vs.~non-salient features by examining the trained network's response to perturbations to task-irrelevant portions of the input (e.g., the right digit in Fig.~\ref{fig:motivation}, an image background in the COCO dataset in \Cref{sec:experiments}, etc.).  \citet{hbar} showed that HSIC regularization terms promote feature invariance and improve robustness even without adversarial training; we extend their analysis by integrating HSIC regularization (Eq.~\eqref{eq: hsic loss}) with salient space isolation (\Cref{eq: acdc loss}), which structurally separates class-discriminative and redundant information in the representation space. 
%
To do so, we make the following assumptions on the kernel families used in the regularization terms \Cref{eq: hsic loss}:
%
%
%
%
\begin{assumption}[Kernel Function Boundedness and Universality]\label{asm: bounded KF}
Let $K_x:\mX \times \mX \to \mathbb{R}^{d \times d}$ and $K_z: \mZ \times \mZ \to \mathbb{R}^{k \times k}$ be continuous positive-definite kernels defined on compact metric spaces $\mX$ and $\mathcal{Z}$, respectively. Let $\kxspace$ and $\kzspace$ denote their associated RKHSs. We assume that:
\begin{enumerate}
    \item The kernels are \emph{universal} on $\mX$ and $\mathcal{Z}$, i.e., $\kxspace$ is dense in $\mathcal{C}(\mX, \mathbb{R}^d)$ and $\kzspace$ is dense in $\mathcal{C}(\mathcal{Z}, \mathbb{R}^k)$ under the supremum norm topology;
    \item All functions in these RKHSs are uniformly bounded in the pointwise $2$-norm, that is,
    \begin{equation}
        K_\kxspace := \sup_{f\in \kxspace} \|f\|_{\infty,2} < \infty, 
        \quad \text{and} \quad 
        K_{\kzspace} := \sup_{g\in \kzspace} \|g\|_{\infty,2} < \infty,
    \end{equation}
    where $\|f\|_{\infty,2} := \sup_{\vx \in \mX} \|f(\vx)\|_2$.
\end{enumerate}
\end{assumption}
Many widely used kernels are known to be universal on compact subsets of \( \mathbb{R}^d \), including the Gaussian (RBF), Laplacian, and Matérn kernels~\citep{steinwart2001influence, sriperumbudur2011universality}. Universality ensures that the RKHS is rich enough to approximate any continuous function on the domain, while boundedness holds automatically on compact input spaces when the kernel is continuous. These properties collectively justify the use of kernel-based function classes for comparing or aligning with neural network outputs, particularly when both are assumed to operate over compact, bounded input spaces. 

As is common when modeling bounded inputs \citep{maatouk2024truncated,ray2020efficient}, we use the truncated multivariate normal (tMVN) distribution: 
$\vx \sim \mathcal{N}_R(0, \sigma^2 I_d),$ with density
$\tilde{p}(\vx) = \frac{1}{C} \exp\left(-\frac{\|\vx\|^2}{2\sigma^2}\right) \cdot \mathbf{1}_{\|\vx\| \le R},$
where $R > 0$ denotes the truncation radius, $\sigma^2 > 0$ is the variance parameter, and $C$ is the normalization constant.
%
Our main robustness guarantee shows that the sensitivity of the model prediction is controlled by (a) the dimensionality of the salient space and (b) the HSIC between the inputs and salient representations. 

\begin{theorem}[HSIC-Based Robustness Bound]
\label{thm:hsic-robustness}
Let $\vx$ be sampled from a tMVN distribution $\vx \sim \mathcal{N}_R(0, \sigma^2 I_d)$, and let the neural network  $h_\vt : \mathbb{R}^d \to \mathbb{R}^k$ be differentiable almost everywhere with an $L$-Lipschitz encoder $f_\psi$ and a bounded linear output layer with $\|\vW\|_\infty \leq B$. Suppose the RKHSs associated with $K_x$ and $K_z$ satisfy Assumption~\ref{asm: bounded KF}, with kernel sup-norm bounds $K_{\mathcal{F}}, K_{\mathcal{G}}$. 
Let $s\coloneqq\|\vM_s\|_0$ be the count of non-zero entries of the salient mask. Then, for all perturbation maps $ \delta: \mathbb{R}^d \to \mathbb{R}^d$ such that  $ \|\delta(\vx)\|_2 \le r$ for all ${\vx\in\mathcal{X}}$,
 the expected output deviation satisfies
\begin{align}
\mathbb{E}_{\vx} \left[ \| h_\vt(\vx + \delta(\vx)) - h_\vt(\vx) \|_2 \right]
\leq \frac{r R B \sqrt{k s}(L R + \|f_\vp(0)\|_2)}{\sigma^2 K_{\mathcal{F}} K_{\mathcal{G}}} \cdot \operatorname{HSIC}(\vx, \vz_s) +o(r),\label{eq:bound}
\end{align}
where $\vz_s := \vM_s f_\vp(\vx)$ is the salient representation and $\vt=\{\vW, \vp\}$ denotes the collection of neural network parameters.
\end{theorem}


The proof of \Cref{thm:hsic-robustness} can be found in \Cref{app: pf of thm}. Intuitively, with stronger information compression imposed by minimizing both the HSIC term (thereby reducing $\operatorname{HSIC}(\vx, \vz_s))$ and the masks (thereby reducing the salient mask support $s\coloneqq\|\vM_s\|_0$), the model is forced to rely only on salient features: \Cref{thm:hsic-robustness} suggests that, in this case, the perturbation is more likely to end up in the non-salient space, and the majority of the attack does not contribute to the change of the output of the neural network, as $\| h_\vt(\vx + \delta(\vx)) - h_\vt(\vx)\|$ stays small for any  perturbation map $\delta(\cdot)$ with $\|\delta(\vx)\|_2 \le r$.  
%
%
From a technical perspective, our theorem differs from \citet{hbar} in two aspects. First, we sharpen the dependence of the upper bound on the power of the perturbation; this allows us to explicitly link it to the dimension of the salient mask  $s$.  
Second, we extend their binary classification framework (i.e., $k=1$ with one output value) to multi-class classification (with arbitrary $k$) to cover a wider range of classification models. 


Additionally, we can quantify how perturbations will trigger prediction changes from the volume of the entire input domain. In particular, we define
 the salient-active region as
\(
\mathcal{X}_s(\epsilon) := \left\{ \vx \in \mathcal{X}_R : \| \nabla_\vx h_\vt(\vx) \|_F > \epsilon \right\}.
\)
Under the tMVN distribution, the probability that an input falls into this region equals its measure:
\(
\mu\left( \mathcal{X}_s(\epsilon) \right) = \mathbb{P}\left( \| \nabla_\vx h_\vt(\vx) \|_F > \epsilon \right).
\) We can bound this probability as follows:
\begin{corollary}[Salient Region Volume Bound via HSIC]
\label{lem:salient-volume}
Under the same assumptions of \Cref{thm:hsic-robustness}, for any threshold $\epsilon > 0$, the probability that a random input falls into the salient-active input region is upper bounded by
\begin{equation}
\mathbb{P}_{\vx}(\| \nabla_\vx h_\vt(\vx) \|_F >\epsilon)
\le \frac{1}{\epsilon} \cdot \left( \frac{R B \sqrt{k s}(L R + \|f_\vp(0)\|_2)}{\sigma^2 K_\mathcal{F} K_\mathcal{G}} \cdot \operatorname{HSIC}(\vx, \vz_s) + o(1) \right).
\end{equation}
\end{corollary}
Thus, \Cref{lem:salient-volume} implies that the volume of the salient-active region is tightly controlled by the dimensionality of the salient space and the HSIC between inputs and the salient representation. The proof is in \Cref{app: pf of lm}.

\section{Experiments}
\label{sec:experiments}

In this section, we describe our datasets, comparison methods, experiment setup, and performance metrics. Additional runtime experiments, implementation details, and hyperparameter tuning and configuration protocols provided in the Appendix~\ref{app:reproducibility}. Our code is publicly available at \url{https://github.com/neu-spiral/H-SPLID}.
\begin{table}[t]
\centering
\resizebox{\textwidth}{!}{
\begin{tabular}{p{5.5cm}ccp{5.5cm}c}
\textbf{Dataset} & \textbf{Encoder Model $f_\psi$} & \textbf{Input size} & \textbf{Perturbation type} & \textbf{\# Categories} \\
\midrule
C-MNIST  & LeNet-3 & $1 \times 64 \times 64$ & PGD attack on right digit \citep{pgd}& 10 \\
COCO subset (bear, elephant, giraffe, zebra) & ResNet-18 & $3 \times 224 \times 224$ & PGD and AA attack (block, background, full) \citep{autoattack} & 4 \\
ISIC-2017 (nevus, melanoma, seborrheic keratosis) & ResNet-50 & $3 \times 224 \times 224$ & Real-world corruptions (brightness, defocus, occlusion) \citep{hendrycks2018benchmarking} & 3 \\
ImageNet-9  & ResNet-50 & $3 \times 224 \times 224$ & Background manipulation and removal  & 368 \\
CounterAnimal (Common vs. Counter) & ResNet-50 & $3 \times 224 \times 224$ & Counterfactual backgrounds  & 45 \\
\bottomrule
\end{tabular}}
\smallskip
\caption{\textbf{Datasets and corresponding models}. “Perturbation type” summarizes how backgrounds/contexts are manipulated in our evaluations.}
\label{tab:dataset_models}
\end{table}

\footnotetext{CounterAnimal is a benchmark split (Common/Counter) over multiple species; we follow its predefined taxonomy and report performance across the two splits rather than a fixed class count.}

\begin{figure*}[t]
  \centering
  \begin{subfigure}[t][1.4in][c]{0.48\textwidth}
    \centering
    \includegraphics[height=1.2in]{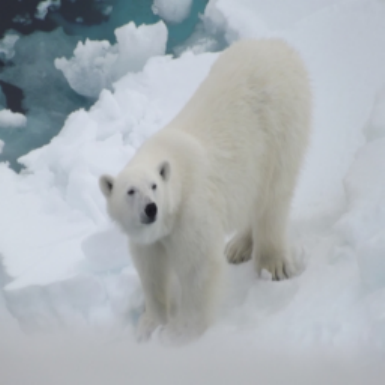}%
    \includegraphics[height=1.2in]{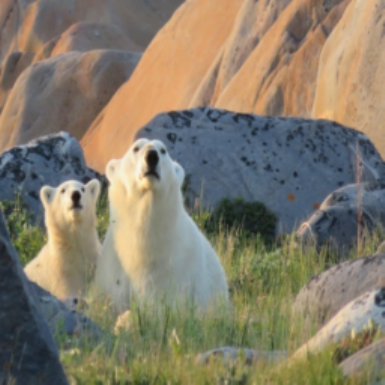}  
    \caption{}
    \label{fig:counteranimal}
  \end{subfigure}
  \hfill
  \begin{subfigure}[t][1.4in][c]{0.48\textwidth}
    \centering
    \includegraphics[height=0.6in]{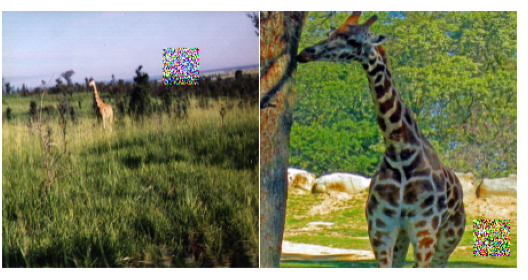}%
    \includegraphics[height=0.6in]{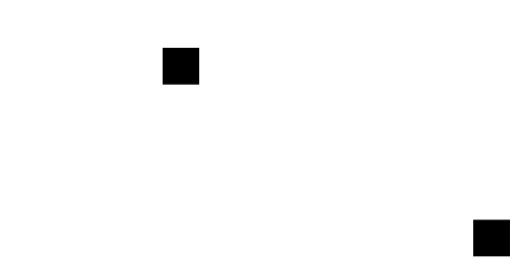}\\
    \includegraphics[height=0.6in]{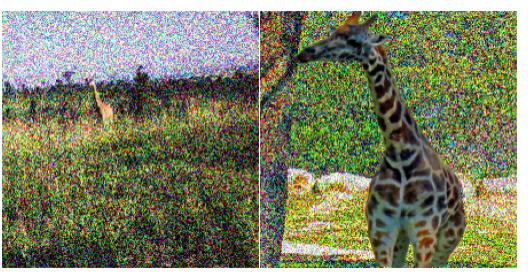}%
    \includegraphics[height=0.6in]{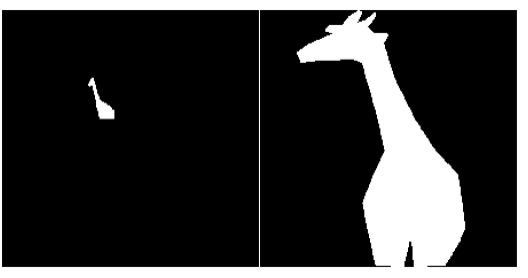}
    \caption{}
    \label{fig:attack_types}
  \end{subfigure}
  \caption{%
    (a) \textbf{Samples from the CounterAnimal dataset.} Common set (left), counter set (right). (b) \textbf{Adversarial attacks on non-salient features.} We attack the non-salient background of COCO images (left) given their corresponding block (upper right) or background mask (lower right) to test whether models successfully learned salient features.
  }
  \label{fig:animals_combined}
\end{figure*}

\paragraph{Datasets and Encoder Models}
We evaluate \algname{} on synthetic and natural image benchmarks, spanning five datasets and three architectures (Table~\ref{tab:dataset_models}). 
We create a synthetic Concatenated-MNIST (C-MNIST) dataset (see Fig.~\ref{fig:motivation}) by concatenating two MNIST digits with the left digit as the class label. We use LeNet-3~\citep{mnist,lenet3} as an encoder.
We construct a four-class subset of COCO~\citep{coco} (\emph{bear}, \emph{elephant}, \emph{giraffe}, \emph{zebra}), coupled  with a ResNet-18~\citep{resnet} encoder. 
ISIC-2017 is a medical imaging dataset~\citep{isic2018skin}. ImageNet-9 (IN-9) \citep{imagenet_9}, encompasses 368 classes from ImageNet-1K instantiated in three variants: \emph{Original} images, a \emph{MixedRand} variant in which object foregrounds are put onto random-class backgrounds, and an \emph{Only-FG} variant with backgrounds entirely removed (See Figure~\ref{fig:imagenet9}). CounterAnimal (CA) \citep{counteranimal},  splits iNaturalist wildlife photos into a Common set (exhibiting typical backgrounds) and a Counter set (featuring atypical yet plausible backgrounds) (see Fig.~\ref{fig:counteranimal}). We use ResNet-50 as an encoder for ISIC-2017 and ImageNet derived datasets.

\paragraph{Comparison Methods.}
\label{subsec:comparison_methods}

To demonstrate how \algname{} focuses on task-relevant characteristics, we compare it with several methods for feature selection and weight regularization, including \emph{weight decay} (\(L_2\) regularization) \citep{weight_decay}, \emph{\(L_1\)} regularization \citep{lasso}, \emph{Group-Lasso} regularization \citep{group_lasso}, and two activation-sparsity variants -- one applying \emph{\(L_1\)} penalty to the penultimate layer's activations and another applying a \emph{Group-Lasso} penalty across those activations to promote instance-level sparsity. Finally, we compare against HBaR \citep{hbar} under non-adversarial training, and a \emph{vanilla} baseline trained only with cross-entropy loss. Appendix~\ref{app:reproducibility} provides further details on training and implementation.

We train \algname{} and all comparison methods exclusively on clean data without employing adversarial attacks or having access to saliency masks. In all datasets, we employ a 80-20 validation split for tuning, and use held-out test sets for final evaluation. Following prior art~\citep{hsic_bottleneck, Yamada2012HighDimensionalFS, xu_nocco, hbar}, we use the Normalized Cross Covariance Operator \citep{fukumizu_cond} to get a scale-insensitive HSIC penalty.  All methods are evaluated using clean test accuracy (over three seeds) and robust test accuracy under different attacks, described below. 

\paragraph{Testset Attacks.}
\label{subsec:experiment_setup}
Methods are evaluated w.r.t. a broad array of attacks on non-salient features at test-time. On C-MNIST, we evaluate predictive performance against a PGD attack on the (non-salient) right digit.  For COCO experiments, we pretrain a ResNet-18 \citep{resnet} from random initialization for 100 epochs with cross-entropy, followed by 200 epochs of method-specific training before evaluating on the held-out test set. We test PGD and AA in three ways: random blocks of pixels in the background, pixel perturbations in the background, and full-image attacks. On ISIC-2017, we use a ResNet-50 pretrained on ImageNet for feature extraction, train a three-class head for 50 epochs, and then run 50 epochs of method-specific training. We test robustness under real-world corruptions (brightness, defocus blur, and snow/occlusion from the corruptions benchmark~\citep{hendrycks2018benchmarking}) applied to non-salient regions (non-lesion pixels). We use IN-9 and CA for transfer learning experiments as follows. First, we train a ResNet-50 \citep{resnet} initialized from ImageNet-1K pretrained weights (TorchVision \citep{torchvision}) for 20 epochs of method-specific training on ImageNet-1K. Then, we test the model on the IN-9 (the original IN-9 and its MixedRand and Only-FG variants) and also on CA (CA-Common and CA-Counter) evaluation sets (see Table~\ref{tab:transfer_experiments}).
 
PGD is implemented via 10 iterations with a step size $\alpha=0.0156$ and AutoAttack  \citep{autoattack} is implemented using the rand ensemble. Attacks per block (Block Atk., see  Fig.~\ref{fig:attack_types}) are confined to a single randomly placed block in the background, with size $\tfrac{1}{4} \times \tfrac{1}{4}$ of the image dimensions. Attacks restricted to background pixels (Block Atk., Background Atk., see  Fig.~ \ref{fig:attack_types}) use saliency masks,  which are available for COCO and ISIC-2017. Full attacks (Full Atk.) are across the entire image. PGD and AA Attacks are conducted over a range of $\epsilon$ levels, with each configuration repeated across five random seeds.
 Additional implementation details and hyperparameter settings are provided in Appendix~\ref{app:reproducibility}.
     
\label{subsec:metrics}




\setlength{\tabcolsep}{2pt}
\begin{table}[t]
\begin{center}
\begin{tabular}{l >{\hspace{2pt}}l@{\hspace{2pt}} cccccccccc}
& & \textbf{No Atk.} & \multicolumn{3}{c}{\textbf{Block Atk.}} & \multicolumn{3}{c}{\textbf{Background Atk.}} & \multicolumn{2}{c}{\textbf{Full Atk.}} \\
\cmidrule(rl){3-3} \cmidrule(rl){4-6} \cmidrule(rl){7-9}  \cmidrule(rl){10-11}
 & Comp. & $\epsilon=0$ & $\frac{25}{255} \rule{0pt}{2.5ex} \rule[-1.5ex]{0pt}{1ex}$ & $\frac{128}{255}$ & $\frac{255}{255}$ & $\frac{1}{255}$ & $\frac{2}{255}$ & $\frac{3}{255}$ & $\frac{1}{255}$ & $\frac{2}{255}$\\
\hline
\multirow{8}{0.5em}{\rotatebox{90}{PGD}}
 & Va. & \textbf{98.1±0.4} & 56.3±0.6 & 51.3±1.3 & 55.1±0.5 & 75.2±0.2 & 56.6±0.2 & 34.4±0.1 & 55.9±0.2 & 34.2±0.3 \\
 & WD & 94.3±0.7 & 43.9±0.5 & 57.2±0.4 & 76.9±0.8 & 76.3±0.0 & 59.9±0.3 & 43.0±0.1 & 57.8±0.1 & 40.7±0.2 \\
 & GLA & 97.1±0.6 & 60.4±1.2 & \textbf{70.4±0.8} & \textbf{78.8±0.8} & 75.1±0.2 & 57.4±0.3 & 35.3±0.3 & 57.5±0.3 & 37.3±0.3 \\
 & GLW & 92.6±1.1 & 45.4±0.5 & 47.2±0.5 & 58.0±0.6 & 72.6±0.0 & 58.4±0.0 & 42.9±0.1 & 54.9±0.1 & 41.2±0.2 \\
 & LSA & 97.1±0.4 & 57.3±1.2 & 63.5±0.5 & 71.9±0.6 & 71.2±0.1 & 54.1±0.3 & 33.7±0.2 & 51.7±0.2 & 35.0±0.4 \\
 & LSW & 96.0±0.6 & 43.2±0.5 & 42.5±0.8 & 57.0±0.4 & 73.5±0.0 & 55.6±0.1 & 37.0±0.1 & 53.1±0.1 & 33.5±0.1 \\
 & HBaR & 97.1±0.4 & 57.4±1.3 & 54.0±1.1 & 67.0±1.2 & 77.9±0.1 & 60.2±0.3 & 39.9±0.3 & 62.4±0.2 & 41.9±0.3 \\
 \rowcolor{lightblue} & Ours & 97.9±0.3 & \textbf{71.9±0.7} & 68.5±0.5 & 72.3±0.4 & \textbf{78.0±0.1} & \textbf{68.9±0.5} & \textbf{57.5±0.2} & \textbf{66.5±0.1} & \textbf{58.9±0.4} \\
\hline
\multirow{8}{0.5em}{\rotatebox{90}{AA}} 
 & Va. & \textbf{98.1±0.4} & 42.8±0.2 & 21.0±0.8 & 19.9±0.6 & 66.8±0.1 & 41.6±0.1 & 26.4±0.2 & 45.5±0.1 & 20.9±0.1 \\
 & WD & 94.3±0.7 & 38.5±0.8 & 23.0±0.7 & 22.5±0.6 & 74.1±0.0 & 53.4±0.1 & 38.7±0.1 & \underline{53.5±0.0} & 29.3±0.0 \\
 & GLA & 97.1±0.6 & 43.3±1.0 & 27.2±1.0 & 28.9±1.2 & 64.6±0.1 & 39.8±0.1 & 26.8±0.2 & 45.5±0.0 & 21.0±0.1 \\
 & GLW & 92.6±1.1 & 41.3±0.8 & 24.6±0.8 & 23.1±0.6 & 71.1±0.1 & 52.0±0.0 & 38.8±0.1 & 52.1±0.0 & 29.5±0.0 \\
 & LSA & 97.1±0.4 & 40.9±0.7 & 25.5±0.6 & 25.7±0.6 & 62.6±0.1 & 39.5±0.1 & 25.3±0.2 & 42.4±0.1 & 20.3±0.1 \\
 & LSW & 96.0±0.6 & 37.5±0.6 & 20.6±0.7 & 20.5±0.5 & 69.7±0.0 & 47.3±0.1 & 31.8±0.2 & 47.3±0.0 & 20.5±0.1 \\
 & HBaR & 97.1±0.4 & 39.5±0.9 & 21.4±0.8 & 29.2±0.4 & 70.1±0.1 & 45.4±0.1 & 31.4±0.3 & 50.4±0.1 & 25.2±0.2 \\
 \rowcolor{lightblue} & Ours & 97.9±0.3 & \textbf{62.1±0.4} & \textbf{48.8±0.3} & \textbf{48.3±0.5} & \textbf{74.2±0.0} & \textbf{59.6±0.2} & \textbf{52.6±0.1} & \textbf{60.4±0.1} & \textbf{48.8±0.2} \\
\hline
\end{tabular}
\end{center}
\smallskip
\caption{\textbf{Measuring saliency with adversarial attacks on COCO}. 
\algname{} (Ours) improves robustness to adversarial attacks compared to most baselines, with the largest performance gains observed under stronger background-targeted attacks (two middle columns). Here, AA denotes AutoAttack, while PGD denotes Projected Gradient Descent. The attack magnitude $\epsilon$ is indicated using ratios of pixel values, with the strongest attack being $\frac{255}{255}$. All models are trained without adversarial data. Va., WD, LSA, LSW, GLA, and GLW denote Vanilla, Weight Decay, \emph{\(L_1\)} Sparse Activations, \emph{\(L_1\)} Sparse Weights, Group-Lasso Activations, and Group-Lasso Weights, respectively.
}
\label{tab:adversarial_robustness}
\end{table}

\subsection{Results}

\paragraph{Controlled attack Benchmark COCO.}
We quantitatively demonstrate the ability of \algname{} to learn salient features on the four-class COCO benchmark by evaluating adversarial robustness under block, background, and full‐image perturbations. As shown in Table \ref{tab:adversarial_robustness}, \algname{} achieves 57.5\% under background-only PGD attacks at $\epsilon=3/255$, with the closest competitor attaining 43.0\%. Even when attacks span the entire image, \algname{} sustains 58.9\% accuracy under a PGD attack with $\epsilon=2/255$, surpassing the 34.2\% of the vanilla network and 41.9\% of the best performing competitor. Against the stronger AutoAttack ensemble, \algname{} consistently outperforms all baselines in robustness to adversarial perturbations. 

These results show that explicitly decomposing latent features into salient and non‐salient subspaces delivers substantial robustness gains, with the most pronounced improvements occurring under background‐only perturbations, validating that \algname{} effectively isolates redundant information. Moreover, robustness gains are achieved without any adversarial training, demonstrating that \algname{}'s latent decomposition strategy yields inherently saliency preserving representations.

\paragraph{Medical imaging Benchmark ISIC-2017.}
To further assess domain generality and robustness, we evaluate \algname{} on the ISIC-2017 skin lesion classification dataset~\citep{isic2018skin} (three classes: nevus, melanoma, seborrheic keratosis). We perturb only non-salient regions (e.g., non-lesion pixels) and adopt three real-world corruptions from the corruptions benchmark~\citep{hendrycks2018benchmarking}: \emph{brightness} (lighting), \emph{defocus blur} (blur), and \emph{snow} (which effectively occludes small patches; we report it as \emph{occlusion}). Results are averaged over 10 random seeds.

\setlength{\tabcolsep}{2pt}
\begin{table}[t]
\begin{center}
\begin{tabular}{l >{\hspace{2pt}}l@{\hspace{2pt}} cccc}
& & \textbf{No Perturb.} & \textbf{Brightness} & \textbf{Defocus} & \textbf{Occlusion} \\
\hline
 & Va.  & 75.45$\pm$0.986 & 66.43$\pm$2.527 & 63.77$\pm$2.388 & 62.87$\pm$3.081 \\
 & WD   & 75.63$\pm$1.545 & 67.53$\pm$2.980 & 64.57$\pm$2.295 & 63.55$\pm$3.851 \\
 & LSA  & 75.62$\pm$1.211 & 66.50$\pm$2.171 & 64.33$\pm$1.653 & 62.27$\pm$4.256 \\
 & LSW  & 75.30$\pm$1.040 & 66.13$\pm$2.432 & 63.22$\pm$2.506 & 62.70$\pm$3.810 \\
 & GLA  & 75.38$\pm$1.383 & 66.50$\pm$3.136 & 61.32$\pm$4.501 & 62.68$\pm$2.969 \\
 & GLW  & 70.65$\pm$4.118 & 60.23$\pm$6.698 & 58.63$\pm$9.564 & 60.62$\pm$5.697 \\
 & HBaR & 75.90$\pm$0.844 & 68.70$\pm$1.942 & 65.62$\pm$2.058 & 66.18$\pm$3.013 \\
\rowcolor{lightblue}
 & Ours & \textbf{76.78$\pm$0.778} & \textbf{70.00$\pm$1.619} & \textbf{68.38$\pm$1.376} & \textbf{69.50$\pm$1.716} \\
\hline
\end{tabular}
\end{center}
\smallskip
\caption{\textbf{Measuring saliency with real-world perturbations on ISIC-2017}. 
\algname{} (Ours) achieves the best robustness across lighting (brightness), blur (defocus), and occlusion (snow) when perturbations are restricted to non-salient regions. All models are trained without adversarial data. Va., WD, LSA, LSW, GLA, and GLW denote Vanilla, Weight Decay, \emph{\(L_1\)} Sparse Activations, \emph{\(L_1\)} Sparse Weights, Group-Lasso Activations, and Group-Lasso Weights, respectively.}
\label{tab:isic_robustness}
\end{table}

These medical imaging results mirror our COCO findings: explicitly separating salient from non-salient latents confers consistent robustness gains under realistic, non-adversarial corruptions, especially when perturbations target only non-salient regions. Together with Table~\ref{tab:adversarial_robustness}, this strengthens the evidence that \algname{} learns saliency-preserving representations that generalize beyond natural images to specialized clinical domains.

\paragraph{Saliency Benchmarks.}
\label{subsec:transfer}

We measure the saliency of our model on the ImageNet-9 and CounterAnimal benchmarks. In Table \ref{tab:transfer_experiments}, \algname{} attains the highest accuracy on the IN‐9 test set (76.7\%), outperforming the vanilla baseline by 2.7\% and exceeding the next best regularization method by over 1\%. When the backgrounds are entirely removed (Only‐FG), \algname{} once again surpasses all methods with a 64.5\% test accuracy, demonstrating its ability to distill object‐centric features. On the more challenging MixedRand variant, it achieves a 59.5\% test accuracy, a substantial 3.1\% gain over the strongest baseline. On the CA Common set, which preserves typical contextual correlations, \algname{} matches the top performing method (80.3\% vs. 80.7\%). Finally, on the CA Counter set of atypical contexts, it surpasses all competitors with a 60.6\% test accuracy, a 2.1\% improvement over the HBaR model. 
\begin{table}[t]
\centering
\begin{tabular}{lccccc}
Method                  & IN-9        & Only-FG      & MixedRand     & CA-Common     & CA-Counter    \\
\hline
Vanilla                 & 74.0        & 60.5         & 51.2          & 78.3          & 58.4 \\
Weight Decay       & 72.6        & 58.4         & 51.2          & 77.3          & 54.4           \\
Group Lasso Activations  & \underline{75.3} & \underline{63.8} & 55.7          & 79.9          & 58.3           \\
Group Lasso Weights     & 73.0        & 60.0         & 50.6          & 78.3          & 57.1           \\
\emph{\(L_1\)} Sparse Activations    & 74.8        & 62.9         & \underline{56.4} & \textbf{80.7} & 58.1           \\
\emph{\(L_1\)} Sparse Weights       & 73.7        & 61.3         & 51.9          & 78.1          & 54.7           \\
HBaR               & 73.6        & 63.3         & 53.8          & 79.3          & \underline{58.5}           \\
\rowcolor{lightblue}
\algname{}              & \textbf{76.7} & \textbf{64.5} & \textbf{59.5} & \underline{80.3} & \textbf{60.6} \\
\hline
\end{tabular}
\smallskip
\caption{\textbf{Transfer accuracy on ImageNet-9 and CounterAnimal saliency benchmarks} 
\algname{} achieves the highest accuracy on the most challenging splits (MixedRand and CA-Counter), demonstrating the robust transferability of its learned representations.}
\label{tab:transfer_experiments}
\end{table}
The consistent performance across original, background‐altered and contextually shifted datasets demonstrates that the explicit separation of salient and non‐salient subspaces in \algname{} yields representations that transfer more robustly to new tasks and real-world perturbations.



    

\paragraph{Loss Term Ablations.}
\label{subsec:ablations}
We ablate the loss components according to their conceptual grouping, namely  cross-entropy ($\mathcal{L}_{ce}$), cross-entropy loss with space separation ($\mathcal{L}_{ce} + \mathcal{L}_{s} + \mathcal{L}_{n}$), cross-entropy loss with $\mathrm{HSIC}$ components ($\mathcal{L}_{ce} + \mathrm{HSIC}(\vX,\vZ_s) + \mathrm{HSIC}(\vY,\vZ_n)$) and the full \algname{} loss ($\mathcal{L}$). The mask computation (\Cref{sec:altopt}) remains unchanged across all ablations (the difference is whether the clustering loss terms $\mathcal{L}_{s} + \mathcal{L}_{n}$ are optimized). All ablations are performed starting from the best‐performing COCO model (Appendix \ref{app:hyperparameters}) by removing individual components of the full \(\mathcal{L}\) objective. Each loss combination was independently tuned to achieve its best performance. As shown in Table~\ref{tab:ablation_robustness}, simply using the cross-entropy loss yields poor background robustness (33.75\%). Adding the \(\mathcal{L}_s\) and \(\mathcal{L}_n\) terms or the HSIC penalties improves robustness to approximately 43-46\%, while maintaining clean accuracy above 96\%. The complete objective results in the best robust performance (57.12\%) while maintaining competitive clean accuracy (97.59\%). We further assess the sensitivity of \algname{} to its hyperparameters in 
Appendix~\ref{app:sensitivity}.

\setlength{\tabcolsep}{2pt}
\begin{table}[t]
\begin{center}
\begin{tabular}{l ccc}
\textbf{Method} & \textbf{No Atk.} & \textbf{Background Atk.} & \textbf{Full Atk.} \\
\hline
$\lambda_{ce}\mathcal{L}_{ce}$ & \textbf{98.30} & 33.75±0.1 & 35.29±0.5 \\
$\lambda_{ce}\mathcal{L}_{ce}$ + $\lambda_{s}\mathcal{L}_{s}$ + $\lambda_{n}\mathcal{L}_{n}$ & 97.52 & 43.69±0.2 & 44.12±0.3 \\
$\lambda_{ce}\mathcal{L}_{ce}$ + $\rho_{s}\mathrm{HSIC}(\vX,\vZ_s)$ + $\rho_{n}\mathrm{HSIC}(\vY,\vZ_n)$ & 96.74 & 42.71±0.3 & 45.87±0.8 \\
\rowcolor{lightblue}
\algname{} (full $\mathcal{L}$) & 97.59 & \textbf{57.12±0.3} & \textbf{58.44±0.2} \\
\hline
\end{tabular}
\end{center}
\smallskip
\caption{\textbf{Ablation of loss terms on COCO.} Accuracy under no attack, Background Attack ($\epsilon=3/255$), and Full Attack ($\epsilon=2/255$) using PGD. Attacks run with five random seeds. The complete objective delivers the highest robust performance.}
\label{tab:ablation_robustness}
\end{table}

\section{Limitations \& Conclusion}
\label{sec:conclustion_future_work}

\paragraph{Limitations.} \algname{} assumes the presence of irrelevant information in the input, as well as a sufficiently diverse dataset in which class-specific features occur across varying contexts. If a particular feature always co-occurs with the same context, \algname{} cannot separate salient from non-salient information, since both appear inseparably -- a challenge that would require external knowledge to resolve. Further, we restricted our analysis to image data, where the distinction between salient and non-salient regions is intuitive to humans. Investigating whether similar decompositions apply to other data modalities remains an exciting direction for future work.

\paragraph{Conclusion.} We introduce \algname{}, a novel method for salient feature learning that decomposes the latent space of a neural network into task-relevant and task-irrelevant components during training. Unlike prior work, \algname{} performs supervised feature selection in an end-to-end manner, without relying on external saliency annotations. Our theoretical analysis provides formal insight into how this decomposition promotes compact and informative representations. Empirically, we show that \algname{} learns class-discriminative features and naturally reduces reliance on irrelevant input variations. In future work, we would like to combine \algname{} with self-supervised models such as I-JEPA \citep{ijepa}, with the goal of learning features that generalize better to downstream tasks. Additionally, we plan to explore the decomposition of salient and non-salient spaces in other data modalities, including graphs, text, and multi-modal data.


\begin{ack}
We gratefully acknowledge support from the National Science Foundation through grant CNS-2414652. Further, this work is supported in part by NIH 5U24CA264369-03. We acknowledge the EuroHPC Joint Undertaking for awarding this project access to the MareNostrum supercomputer (hosted at BSC, Spain), MeluXina (operated by LuxProvide, Luxembourg), Deucalion (hosted at the Minho Advanced Computing Center, Portugal), and Discoverer (hosted at Sofia Tech, Bulgaria) through EuroHPC Access allocations.

\end{ack}






\bibliographystyle{abbrvnat}
\bibliography{lib}


\appendix
\newpage
\section{Technical Preliminary}\label{app: bg kn}
\noindent{\bf Hilbert Space.} A Hilbert space is a complete inner product space. More formally, a real or complex vector space \( \mathcal{H} \) is called a Hilbert space if it is equipped with an inner product \( \langle \cdot, \cdot \rangle_\mathcal{H} \) that induces a norm \( \|f\|_{\mathcal{H}} := \sqrt{\langle f, f \rangle_\mathcal{H}} \), under which \( \mathcal{H} \) is complete; that is, every Cauchy sequence in \( \mathcal{H} \) converges to a limit in \( \mathcal{H} \). The inner product structure generalizes the geometric notions of angle and length, while completeness ensures that limits of convergent sequences remain in the space. Examples include \( \mathbb{R}^n \), \( L^2 \) spaces of square-integrable functions, and reproducing kernel Hilbert spaces (RKHSs).


\noindent{\bf Reproducing Kernel Hilbert Space (RKHS).}
Let $ \mathcal{X} $ be a nonempty set. A Hilbert space $ \mathcal{H} \subseteq \mathbb{R}^\mathcal{X} $ is called a \emph{reproducing kernel Hilbert space} if there exists a positive definite kernel $ k : \mathcal{X} \times \mathcal{X} \to \mathbb{R} $ such that, for every $ \vx \in \mathcal{X} $, the function $ k(x, \cdot) \in \mathcal{H} $ and the reproducing property holds: that is, for all $ f \in \mathcal{H} $ and $ x \in \mathcal{X} $,
\[
f(\vx) = \langle f, k(\vx, \cdot) \rangle_{\mathcal{H}}.
\]

\noindent{\bf Hilbert-Schmidt Operator {\citep{gretton2005measuring}}.}
Let $\mathcal{F}$ and $\mathcal{G}$ be separable Hilbert spaces, and let $A : \mathcal{F} \to \mathcal{G}$ be a bounded linear operator.\footnote{Separable Hilbert spaces implies the spaces have a complete orthonormal basis.} Then $A$ is called a Hilbert-Schmidt operator if, for any orthonormal bases $\{f_i\}_{i=1}^\infty \subset \mathcal{F}$ and $\{g_j\}_{j=1}^\infty \subset \mathcal{G}$, the following Hilbert-Schmidt norm is finite:
\begin{equation}
\|A\|_{\mathrm{HS}}^2 := \sum_{i=1}^\infty \sum_{j=1}^\infty  \langle A f_i, g_j \rangle_{\mathcal{G}} ^2 < \infty.
\end{equation}



\noindent{\bf Cross-Covariance Operator.}
Let $\vx \in \mathcal{X}$, $\vz \in \mathcal{Z}$ be random variables and let $\mathcal{F}$ and $\mathcal{G}$ be RKHSs over $\mathcal{X}$ and $\mathcal{Z}$. Then, the \emph{cross-covariance operator} $C_{XZ} : \mathcal{G} \to \mathcal{F}$ is the unique linear operator such that
\begin{equation}
\langle f, C_{XZ} g \rangle_{\mathcal{F}} := \operatorname{Cov}[f(\vx), g(\vz)] = \mathbb{E}\left[ (f(\vx) - \mathbb{E}[f(\vx)])(g(\vz) - \mathbb{E}[g(\vz)]) \right],
\end{equation}

for all $f \in \mathcal{F}$, $g \in \mathcal{G}$.

\begin{proposition}[Covariance Bounded by HSIC {\citep{gretton2005measuring,robustlearning-v119-greenfeld20a}}]
\label{prop:hsic-bounds-sum-cov}
Let $X \in \mX$ and $\vz \in \mZ$ be random variables, and let $\mF$ and $\mG$ be RKHSs on $\mX$ and $\mZ$, respectively. Then the scalar covariance is bounded by the Hilbert-Schmidt Information Criterion, i.e., the Hilbert-Schmidt norms of the cross-covariance operators:
\begin{equation}
\sup_{f \in \mF \\ g \in \mG}
\operatorname{Cov}[f(\vx), g(\vz)] \leq \operatorname{HSIC}_{s}(\vx,\vz)\equiv  \left\| C_{XZ}\right\|_{\mathrm{HS}}.
\end{equation}
\end{proposition}

The above proposition is the HSIC for the scalar value RKHS defined in \citet{gretton2005measuring}. To connect the above scalar value function spaces to vector-value spaces, we use the external direct sum as below.

\noindent{\bf External Direct Sum of Hilbert Spaces.} Let $\mathcal{H}_1, \ldots, \mathcal{H}_k$ be Hilbert spaces. We can then denote the vector-valued Hilbert space $\mathcal{H}$ via the external direct sum as
\begin{equation}\label{eq: direct sum}
\mathcal{H} := \bigoplus_{j=1}^k \mathcal{H}_j := \left\{ (f_1, \ldots, f_k) \mid f_j \in \mathcal{H}_j \right\},
\end{equation}
which is equipped with the inner product $\left\langle (f_1, \ldots, f_k), (g_1, \ldots, g_k) \right\rangle_{\mH} := \sum_{j=1}^k \langle f_j, g_j \rangle_{\mathcal{H}_j}$. The corresponding norm is given by $\|f\|_{\mathcal{H}} := \left( \sum_{j=1}^k \|f_j\|_{\mathcal{H}_j}^2 \right)^{1/2}$,
which makes $\mathcal{H}$ itself a Hilbert space.

Moreover, if we construct the RKHS by the direct sum, i.e., $\mathcal{H} := \bigoplus_{j=1}^k \mathcal{H}_j$, the resulting space $\mH$ is a vector-valued RKHS \citep{carmeli2010}. 

Next, we define the corresponding covariance matrix for a vector-valued RKHS. Let $f: \mX \to \mathbb{R}^k$ and $g: \mZ \to \mathbb{R}^k$ be vector-valued functions, and $(\vx, \vz)$ be random variables jointly distributed over $\mX \times \mZ$. The covariance matrix between $f$ and $g$ is defined as:
\begin{equation}
\operatorname{Cov}[f(\vx), g(\vz)] := \mathbb{E}\left[ (f(\vx) - \mathbb{E}[f(\vx)]) (g(\vz) - \mathbb{E}[g(\vz)])^\top \right] \in \mathbb{R}^{k \times k}.
\end{equation}


{\bf Hilbert-Schmidt Independence Criterion (HSIC)} The HSIC \citep{gretton2005measuring} is a kernel-based measure of dependence. Let $\vx \in \mathcal{X} \subseteq \mathbb{R}^d$ and $\vz \in \mathcal{Z} \subseteq \mathbb{R}^m $ be random variables.  Let also $\mathcal{G} = \bigoplus_{i=1}^k \mathcal{G}_i$ and $\mathcal{F} = \bigoplus_{j=1}^k \mathcal{F}_j$ be vector-valued RKHSs over the input and representation domains with $k$ values, respectively (i.e., direct sums of $k$ scalar RKHSs).  The \emph{cross-covariance operator} $C_{XZ} : \mathcal{G} \to \mathcal{F}$ is the unique linear operator such that
$\langle f, C_{XZ} g \rangle_{\mathcal{F}} := \operatorname{Cov}[f(\vx), g(\vz)] = \mathbb{E}\left[ (f(\vx) - \mathbb{E}[f(\vx)])(g(\vz) - \mathbb{E}[g(\vz)]) \right],$
for all $f \in \mathcal{F}$, $g \in \mathcal{G}$.
The vector-valued HSIC between $\vx$ and $\vz$ is defined as
\begin{align}
\label{def: vvHSIC}
\operatorname{HSIC}(\vx,\vz) := \sum_{j=1}^k \sum_{i=1}^k \| C_{XZ}^{(i,j)} \|_{\mathrm{HS}},
\end{align}
where $C_{XZ}^{(i,j)}$ is the cross-covariance operator between $\mathcal{G}_i$ and $\mathcal{F}_j$, and $\| \cdot \|_{\mathrm{HS}}$ denotes the Hilbert-Schmidt norm. This quantity upper-bounds the scalar covariances \citep{robustlearning-v119-greenfeld20a}:
\[
\sup_{f_j \in \mathcal{F}_j,\, g_i \in \mathcal{G}_i} \operatorname{Cov}[f_j(\vx), g_i(\vz)] \le \| C_{XZ}^{(i,j)} \|_{\mathrm{HS}}.
\]
%
%
We can empirically estimate HSIC within $O(n^{-1})$ accuracy \citep{gretton2005measuring} 
given $n$ i.i.d. samples $\{(x_i, y_i)\}_{i=1}^n$ via:
\begin{align}\label{eq:hhsic}\widehat{\operatorname{HSIC}}(\vX,\vZ)=\sum_{j=1}^k \sum_{i=1}^k (n-1)^{-2} \operatorname{tr}(K_x H K_z H),
\end{align}
where $K_x$ and $K_z $ have elements $K_{x_{(i,j)}}=k_x(\vx_i, \vx_j)$ and $K_{z_{(i,j)}}=k_z(\vz_i, \vz_j)$, while $H=\mathbf{I}-\frac{1}{n} \mathbf{1} \mathbf{1}^\top$ is the centering matrix.


\paragraph{Notation Summary.}
\Cref{tab:notation} summarizes the notation used in the main paper and Appendix.

\begin{table}[!t]
\centering
\caption{Summary of Notation and Terminology}
\begin{tabular}{ll}
\toprule
\textbf{Symbol} & \textbf{Description} \\
\midrule
\multicolumn{2}{l}{\textbf{Domains and Variables}} \\
\midrule
$\mathcal{X} \subseteq \mathbb{R}^d$ & Input domain (bounded subset of $\mathbb{R}^d$) \\
$\mathcal{R} \subseteq \mathbb{R}^m$ & Representation domain (output of encoder) \\
$\mathcal{Z} \subseteq \mathbb{R}^m$ & Salient representation domain (output of encoder) \\
$\mathbb{R}^k$ & Output/logit space \\
$X \sim \mathcal{N}_R(0, \sigma^2 I_d)$ & Truncated multivariate normal on ball of radius $R$ \\
$Z := h_\vt(\vx)$ & Output representation of the network \\
\midrule
\multicolumn{2}{l}{\textbf{Functions and Network Components}} \\
\midrule
$\pi_i(\vx) := x_i$ & $i$-th coordinate projection function \\
$f_\vp : \mathcal{X} \to \mathcal{R}$ & $L$-Lipschitz encoder network \\
$\vM \in \mathbb{R}^{m \times m}$ & Diagonal binary mask matrix (entries in $\{0,1\}$) \\
$s := \operatorname{tr}(\vM)$ & Number of active (nonzero) dimensions in the mask \\
$\vW \in \mathbb{R}^{k \times m}$ & Linear weight matrix after masking \\
$g_\vW : \mathcal{Z} \to \mathbb{R}^k$ & Final representation-domain function (e.g., linear layer) \\
$h_\vt := g_\vW \circ f_\vp$ & Full neural network from input to output \\
\midrule
\multicolumn{2}{l}{\textbf{Norms and Constants}} \\
\midrule
$\|\cdot\|_2$ & Euclidean (vector) norm \\
$\|\cdot\|_F$ & Frobenius norm for matrices \\
$\|\cdot\|_\infty$ & Maximum absolute value across components (for vectors) \\
$\|\cdot\|_{\infty,2}$ & Supremum of 2-norm: $\|f\|_{\infty,2} := \sup_{\vx \in \mathcal{X}} \|f(\vx)\|_2$ \\
$B := \|\vW\|_\infty$ & Max row-wise $\ell_1$ norm of the weight matrix $\vW$ \\
$N_\mathcal{X} := R$ & Sup-2-norm bound of input: $\|x\|_2 \le R$ \\
$N_\mathcal{Z} := B \sqrt{k s}(L R + \|f_\vp(0)\|_2)$ & Sup-2-norm bound on $g_\vW \circ f_\vp$ \\
$\|\delta\|_2 \le r$ & Perturbation is bounded by \(r\)\\
\midrule
\multicolumn{2}{l}{\textbf{Kernel and RKHS Quantities}} \\
\midrule
$k_\mathcal{X}$, $k_\mathcal{Z}$ & Universal kernels on input and representation domains \\
$\mathcal{F}$, $\mathcal{G}$ & RKHSs induced by $k_\mathcal{X}$ and $k_\mathcal{Z}$ \\
$K_\mathcal{F}, K_\mathcal{G}$ & Kernel Sup-2-norm bounds\\
$\operatorname{HSIC}(\vx, \vz)$ & Hilbert-Schmidt Independence Criterion between $X$ and $\vz$ \\
\midrule
\multicolumn{2}{l}{\textbf{Loss and Perturbation}} \\
\midrule
$\delta \in \mathbb{R}^d$ & Input perturbation with $\|\delta\|_2 \le r$ \\
$\mathcal{L}(h_\vt(\vx), y)$ & Loss function for prediction and ground-truth $y$ \\
$L_\mathcal{L}$ & Lipschitz constant of the loss in its first argument \\
\bottomrule
\end{tabular}
\label{tab:notation}
\end{table}

\section{\algname{} Pseudocode }
\label{app: psd code}

\Cref{alg:alt-opt} contains pseudo-code for \algname, i.e., the alternating optimization algorithm presented in \Cref{sec:altopt} to solve Problem~\eqref{eq:prob}.
\begin{algorithm}[t]
\caption{Alternating Optimization of \( \vt \) and \( \vM_s \)}
\label{alg:alt-opt}
\begin{algorithmic}[1]
\State \textbf{Input:} Dataset \( \mathcal{D} = \{ (\vx_i, y_i) \}_{i=1}^n \); initial \( \vt^{(0)} \); \( \vM_s^{(0)} = I \);  \(\beta_{step}\in [0,1]\);

\For{epoch \( t = 1 \) to \( T \)}
    \Statex \textbf{Step 1: Update \( \vt \) via SGD (minibatches)}
    \For{each minibatch \( \mathcal{B} \subset \mathcal{D} \)}
        \State Compute latent codes \( \vz_i = f_{\vp}(\vx_i) \) for \( i \in \mathcal{B} \)
        \State Compute minibatch means: \( \boldsymbol{\mu} \), \( \{ \boldsymbol{\mu}_k \}_{k=1}^K \)
        \State Compute loss \( \mathcal{L}(\mathcal{B}; \vt, \vM_s, I - \vM_s) \)
        \State Update \( \vt \leftarrow \vt - \eta \nabla_{\vt} \mathcal{L} \)
    \EndFor

    \Statex \textbf{Step 2: Update \( \vM_s \) via closed-form solution}
    \State Compute \( \vz_i = f_{\vp}(\vx_i) \) for all \( i \in \mathcal{D} \)
    \State Compute dataset means: \( \boldsymbol{\mu} \), \( \{ \boldsymbol{\mu}_k \}_{k=1}^K \)
    \For{each feature dimension \( i = 1 \) to \( m \)}
        \Statex
        \begin{equation*}
        (\vM_s)_{i,i} \leftarrow \beta_{step}(\vM_s)_{i,i}+(1-\beta_{step})
        \frac{
        \lambda_n \sum_{\vz \in \mathcal{D}} (\vz_i - \boldsymbol{\mu}_i)^2
        }{
        \lambda_s \sum_{k=1}^K \sum_{\vz \in C_k} (\vz_i - (\boldsymbol{\mu}_k)_i)^2 +
        \lambda_n \sum_{\vz \in \mathcal{D}} (\vz_i - \boldsymbol{\mu}_i)^2
        }
        \end{equation*}
    \EndFor
\EndFor

\State \textbf{Return:} \( \vt^{(T)} \), \( \vM_s^{(T)} \)
\end{algorithmic}
\end{algorithm}

\section{Proof of Theorem~\ref{thm:hsic-robustness}} 
\label{app: pf of thm}
We next show that the output of a masked neural network is uniformly bounded in sup-norm under standard Lipschitz and compactness conditions. This provides the foundation for connecting the model class to the kernel-bounded spaces introduced above.

\begin{lemma}[Bounded NN with Saliency Space]
\label{lem:mask-sparsity-bound}
Let $\mathcal{X} \subseteq \{ \vx \in \mathbb{R}^d : \; \rVert \vx \lVert_2 \le R \}$ be a bounded input space, and let $f_\vp : \mathcal{X} \to \mathbb{R}^m$ be an $L$-Lipschitz encoder. Consider a network $h_\vt(\vx) := \vW \vM f_\vp(\vx)$ where $\vW \in \mathbb{R}^{k \times m}$ is a linear weight matrix satisfying $\|\vW\|_\infty \le B$, and $\vM \in \mathbb{R}^{m \times m}$ is a diagonal binary mask with at most $s$ nonzero entries. Then, the network output is bounded in sup-norm:
\begin{equation}
\|h_\vt\|_{\infty,2} := \sup_{\vx \in \mathcal{X}} \|h_\vt(\vx)\|_2 \le B \sqrt{k s} (L R + \|f_\vp(\mathbf{0})\|_2).
\end{equation}
\end{lemma}

The proof is deferred to Appendix~\ref{pf:mask-sparsity-bound}. This bound shows that the sparsity level $s$ of the mask plays a direct role in constraining the model's output magnitude, which is essential for robustness.

To connect neural network outputs to kernel-based function spaces, we reparameterize the neural network $h_\vt(\cdot) := \vW \vM_s f_\vp(\cdot)$ by $g_\vW(\vz_s)\equiv \vW \vz_s$. Then, we show how the $g_\vW$ belongs to a bounded subset of $\mathcal{C}(\mathcal{Z}, \mathbb{R}^k)$. 

\begin{corollary}[Bounded Function Spaces]\label{cor: bounded-nn-ball}
By redefining the neural network in \Cref{lem:mask-sparsity-bound}, $g_\vW$ belongs to the following closed ball:
\begin{equation}
\logits \in \mathcal{C}_b^{N_\mathcal{Z}} := \left\{ g \in \mathcal{C}(\mathcal{Z}, \mathbb{R}^k) \,\middle|\, \|g\|_{\infty,2} \le N_\mathcal{Z}\equiv B \sqrt{k s} (L R + \|f_\vp(\mathbf{0})\|_2)\right\},
\end{equation}
where $ \|g\|_{\infty,2} := \sup_{\vz \in \mZ} \|g(\vz)\|_2$ denotes the sup-2-norm.
\end{corollary}

This corollary imposes uniform boundedness of the neural network output values via $g_\vW$ on representations over the compact domain $\mX$, ensuring that the function belongs to a bounded subset of continuous function spaces $\mathcal{C}_b^{N_\mathcal{Z}}$. See \Cref{pf: cor bounded nn ball} for the proof.

Given the RKHSs $\mF$ and $\mG$ in \Cref{asm: bounded KF}, we define the rescaled RKHS spaces $\hat{\mF}$ and $\hat{\mG}$ as
\begin{equation}
\hat{\mF} := \left\{ \frac{N_{\mX}}{K_{\mF}} f : f \in \mF \right\} \quad \textit{and} \quad
\hat{\mG} := \left\{ \frac{N_{\mathcal{Z}}}{K_{\mG}} f : f \in \mG \right\}.
\end{equation}

Thus, we establish the equivalence between the rescaled RKHSs and the bounded continuous function spaces.

\begin{lemma}[Rescaled RKHS Equals $\mathcal{C}_b(\mX, \mathbb{R}^d)$]
\label{lemma:rescaled RKHS equality}
Given \Cref{asm: bounded KF} and the continuous universal kernel $k_x$ therein, its corresponding RKHS $\mF$, and a bounded continuous function space $\mathcal{C}_b(\mX, \mathbb{R}^d)$ such that 
\[
\mathcal{C}_b^{N_\mX} := \left\{ f \in \mathcal{C}(\mX, \mathbb{R}^d) \,\middle|\, \|f\|_{\infty,2} \le N_\mX \right\}, 
\]
then we have the rescaled RKHS space
\begin{equation}
\hat{\mF} := \left\{ \frac{N_{\mX}}{K_{\mF}} f : f \in \mF \right\}
\end{equation}
satisfying
\begin{equation}
\overline{\mF}= \mathcal{C}_b^{N_{\mX}} .
\end{equation}
where $\overline{\mF}\coloneqq \overline{\hat{\mF}}^{\|\cdot\|_{\infty,2}} $ denotes the closure of $\hat{\mF}$ w.r.t. the $||\cdot||_{\infty,2}$. 
\end{lemma}

Similarly, we can show that the rescaled $\mathcal{C}_b^{N_{\mZ}}=\overline{\hat{\mG}}^{\|\cdot\|_{\infty,2}} $.
See \Cref{pf: rescaled RKHS equality} for the proof.

Moreover, as we have two spaces containing functions that are different by a scalar, we are interested in how the supremums of the covariance relate between spaces.

\begin{lemma}[Scaling of Supremum Covariance — Sum Version]
\label{lemma:scaling-sup-cov-sum}
Let $\mF$ and $\mG$ be vector-valued RKHSs over $\mX$ and $\mZ$, respectively. Then, for all $M_{\mF}, M_\mG > 0$ and $\vx \in \mX, \vz_s \in \mZ$, the following holds:
\begin{equation}
\sum_{j=1}^{k} \sum_{i=1}^{d} \sup_{f_j \in \mF_j, g_i \in \mG_i} \operatorname{Cov}[f_j(\vx), g_i(\vz_s)]
= M_{\mF} M_{\mG} 
\sum_{j=1}^{k} \sum_{i=1}^{d} \sup_{\tilde{f}_j \in \hat{\mF}_j, \tilde{g}_i \in \hat{\mG}_i} \operatorname{Cov}[\tilde{f}_j(\vx), \tilde{g}_i(\vz_s)],
\end{equation}
where $\hat{\mF} := \{ \tilde{f} = f / M_{\mF} : f \in \mF \}$ and likewise for $\hat{\mG}$.
\end{lemma}
See \Cref{pf: scaling-sup-cov-sum} for the proof.

Lastly, given the supremum of covariance of the function space containing the neural network, we need to use the variant of Stein's Lemma to bound the gradient of the neural network.

\begin{lemma}[Stein’s Lemma for Scalar-Valued Functions on a Bounded Domain]
\label{lem:bounded-stein-gradient-bound}
Let $\vx$ be sampled from a truncated multivariate normal (tMVN) distribution, i.e., with density $\tilde{p}(\vx) = \frac{1}{C} \exp\left(-\frac{\|\vx\|^2}{2\sigma^2}\right) \mathbf{1}_{\|x\| \leq R},$
supported on the compact domain $\mathcal{X}_R := \{\vx \in \mathbb{R}^d : \|\vx\| \le R\}$, where $C$ is the normalization constant. Let $h : \mathbb{R}^d \to \mathbb{R}$ be a differentiable almost everywhere such that $\mathbb{E}|\partial h(\vx)/\partial x_i| < \infty$ and $|h(\vx)| \le N_{\mathcal{X}}$ for all $x \in \mathcal{X}_R$. Then for all $i \in \{1, \dots, d\}$,
\begin{equation}
\mathbb{E} \left[ \frac{\partial h(\vx)}{\partial x_i} \right]
= \frac{1}{\sigma^2} \operatorname{Cov} \left[ x_i, h(\vx) \right] + \epsilon_i(R),
\end{equation}
where the truncation error term satisfies
\[
|\epsilon_i(R)| \le \frac{N_\mX C_d R^{d-1}}{C} \exp\left( -\frac{R^2}{2\sigma^2} \right),
\]
and $C_d$ is the surface area of the unit sphere in $\mathbb{R}^d$.
\end{lemma}
See \Cref{pf: bounded-stein-gradient-bound} for the proof.

Then, we formally state the proof of \Cref{thm:hsic-robustness}.
\begin{proof}

{\bf Step 1. Continuous function spaces $C_\mX^{N_\mX}$ and $C_\mZ^{N_\mZ}$.}

Let $\pi : \mathcal{X} \to \mathbb{R}^d$ denote the identity map, defined by $\pi(\vx) := \vx$. This vector-valued function can be decomposed into scalar coordinate projections:
\[
\pi_i(\vx) := x_i, \quad \text{for } i = 1, \dots, d.
\]

Since the input domain $\mathcal{X} \subseteq \mathbb{R}^d$ is contained within a Euclidean ball of radius $R$, we have $\|\pi(\vx)\|_2 \le R$ for all $x \in \mathcal{X}$. Therefore, the identity function satisfies:
\[
N_{\mathcal{X}} := \|\pi\|_{\infty,2} = R,
\]
and lies in the vector-valued bounded continuous function space $\pi \in \mathcal{C}_b^{N_{\mathcal{X}}}(\mathcal{X}, \mathbb{R}^d)$. Correspondingly, each coordinate function belongs to the subspace $\pi_i \in C_{b,i}^{N_{\mathcal{X}}}$.

Now consider the function $g_\vW : \mathcal{Z} \to \mathbb{R}^k$ on the representation domain $\mZ$. From \Cref{cor: bounded-nn-ball}, the composed function $g_\vW \circ \vM_s f_\vp$ over $\mX$ satisfies:
\[
N_{\mathcal{Z}} := B \sqrt{k s}(L R + \|f_\vp(0)\|_2).
\]

Similarly, over the representation space $\mZ$, we have $g_\vW \in \mathcal{C}_b^{N_{\mathcal{Z}}}(\mathcal{Z}, \mathbb{R}^k)$, and each scalar component $g_\vW^{(j)} \in C_{b,j}^{N_{\mathcal{Z}}}$.

{\bf Step 2. Equivalence between RKHS and continous function spaces.}

By \Cref{lemma:rescaled RKHS equality}, we can rescale the RKHS $\mathcal{F}$ and $\mathcal{G}$ in \Cref{asm: bounded KF} as
\begin{align}
    \hat{\mF} := \left\{ \frac{N_{\mX}}{K_{\mF}} f : f \in \mF \right\} \quad \textit{and} \quad \hat{\mG} := \left\{ \frac{N_{\mZ}}{K_{\mG}} g : g \in \mG \right\},
\end{align}

so that their closure are equivalent to the bounded continuous function space $C_\mX^{N_\mX}$ and $C_\mZ^{N_\mZ}$ as in step 1. 

According to Lemma~\ref{lemma:scaling-sup-cov-sum}, if we set $M_\mF\coloneqq \frac{K_{\mathcal{F}}}{N_{\mathcal{X}} }, M_\mG\coloneqq \frac{ K_{\mathcal{G}}}{N_{\mathcal{Z}}} $ we relate covariance bounds between the RKHSs ($\mathcal{F}$ and $\mathcal{G}$) and the rescaled RKHSs ($\hat{\mathcal{F}}$ and $\hat{\mathcal{G}}$) through rescaling.

\begin{align}
&\sum_{j=1}^k \sum_{i=1}^d \sup_{f_j\in \mathcal{F}_j,\, g_i \in \mathcal{G}_i} \operatorname{Cov}[f_j(\vx), g_i(\vz_s)] \label{eq: cov trans}\\
&= \frac{K_{\mathcal{F}} K_{\mathcal{G}}}{N_{\mathcal{X}} N_{\mathcal{Z}}} 
\sum_{j=1}^k \sum_{i=1}^d \sup_{\hat{f}_j \in \hat{\mathcal{F}}_j,\, \hat{g}_i \in \hat{\mathcal{G}}_i} \operatorname{Cov}[\hat{f}_j(\vx), \hat{g}_i(\vz_s)]
\tag{\Cref{lemma:scaling-sup-cov-sum}} \\
&= \frac{K_{\mathcal{F}} K_{\mathcal{G}}}{N_{\mathcal{X}} N_{\mathcal{Z}}} 
\sum_{j=1}^k \sum_{i=1}^d \sup_{\hat{f}_j \in \overline{\hat{\mathcal{F}}}_j,\, \hat{g}_i \in \overline{\hat{\mathcal{G}}}_i} \operatorname{Cov}[\hat{f}_j(\vx), \hat{g}_i(\vz_s)]
\tag*{\text{(closure under sup-norm)}} \\
&= \frac{K_{\mathcal{F}} K_{\mathcal{G}}}{N_{\mathcal{X}} N_{\mathcal{Z}}} 
\sum_{j=1}^k \sum_{i=1}^d \sup_{\tilde{f}_j \in C_{b,j}^{N_{\mathcal{X}}},\, \tilde{g}_i \in C_{b,i}^{N_{\mathcal{Z}}}} \operatorname{Cov}[\tilde{f}_j(\vx), \tilde{g}_i(\vz_s)]
\tag{\Cref{lemma:rescaled RKHS equality}},
\end{align}

where the second last line applies closure under the sup-norm (preserving the supremum), and the last line substitutes the equivalent bounded continuous function space by \Cref{lemma:rescaled RKHS equality}.



{\bf Step 3. Bound covariance in continuous function spaces with HSIC.}

Then, based on \Cref{def: vvHSIC}, we obtain:
\begin{equation}\label{eq: vvHSIC}
\sum_{j=1}^k \sum_{i=1}^d \sup_{f_j\in \mF_j, g_i \in \mG_i}  \operatorname{Cov}[f_j(\vx), g_i(\vz_s)] 
\le \operatorname{HSIC}(\vx, \vz_s).
\end{equation}
Combining \cref{eq: cov trans} and \cref{eq: vvHSIC}, we have
\begin{equation}
\sum_{j=1}^k \sum_{i=1}^d \sup_{\tilde{f}_j\in C_{b, j}^{N_\mX}, \hat{g}_i\in C_{b, i}^{N_\mZ}} \operatorname{Cov}[\tilde{f}_j(\vx), \tilde{g}_i(\vz_s)]
\le \frac{N_{\mathcal{X}} N_{\mathcal{Z}}}{K_{\mathcal{F}} K_{\mathcal{G}}} \cdot \operatorname{HSIC}(\vx, \vz_s).
\end{equation}

As shown in Step 1, we have $\pi_i\in C_{b, i}^{N_\mX}, h_\vt^{(j)}\in C_{b, j}^{N_\mZ}$, and the following holds
\begin{equation}\label{eq: HSIC bounds Cov}
\sum_{j=1}^k \sum_{i=1}^d \sup_{\pi_i\in C_{b, i}^{N_\mX}, h_\vt^{(j)}\in C_{b, j}^{N_\mZ}}\operatorname{Cov}[\pi_i(\vx), h^{(j)}_\vt(\vx)] 
\le \frac{N_\mX N_{\mathcal{Z}}}{K_{\mathcal{F}} K_{\mathcal{G}}} \cdot \operatorname{HSIC}(\vx, \vz_s).
\end{equation}

{\bf Step 4. Bound the gradient with covariance.}

By Lemma~\ref{lem:bounded-stein-gradient-bound}, the following holds 
\begin{equation}\label{eq: Cov and der}
 \left|\mathbb{E} \left[ \frac{\partial h^{(j)}_\vt(\vx)}{\partial x_i} \right] 
\right| = \frac{1}{\sigma^2}  \left|\operatorname{Cov}[X_i, h^{(j)}_\vt(\vx)] \right| \leq \frac{1}{\sigma^2}\sup_{\pi_i\in C_{b, i}^{N_\mX}, h_\vt^{(j)}\in C_{b, j}^{N_\mZ}}\operatorname{Cov}[\pi_i(\vx), h^{(j)}_\vt(\vx)] .
\end{equation}

Combining \eqref{eq: HSIC bounds Cov} and \eqref{eq: Cov and der} gives:
\begin{equation}\label{eq: HSIC final bound}
\sum_{j=1}^k \sum_{i=1}^d  \left|\mathbb{E} \left[ \frac{\partial h^{(j)}_\vt(\vx)}{\partial x_i} \right]\right|
\le \frac{1}{\sigma^2} \left( \frac{N_\mX N_{\mathcal{Z}}}{K_{\mathcal{F}} K_{\mathcal{G}}} \cdot \operatorname{HSIC}(\vx, \vz_s) + \epsilon_R \right).
\end{equation}

Consider the  first-order Taylor expansion in a Euclidian ball of radius $r$ around $\vx\in \mathcal{X}$: that is, 
\begin{equation}\label{eq: taylor1}
h_\vt(\vx + \delta) - h_\vt(\vx) = J_{h_\vt}(\vx) \delta + o(r).
\end{equation}
for all $\vx\in \mathcal{X}$ and all $\delta\in\mathbb{R}^d $ s.t. $\|\delta\|_2 \leq r$.

Consider now a measurable perturbation function on $\mathcal{X}$ as $\delta : \mathbb{R}^d \to \mathbb{R}^d$, such that
\begin{align}\label{eq:bound}
\sup_{\vx \in \mathcal{X}} \|\delta(\vx)\|_2 \;\le\; r.
\end{align}

As Eq.~\eqref{eq: taylor1} holds for all $\vx,\delta$ pairs, for all $\vx\in \mathcal{X}$, we have that:
\begin{equation}\label{eq: taylor}
h_\vt(\vx + \delta(\vx)) - h_\vt(\vx) = J_{h_\vt}(\vx) \delta(\vx) + o(r).
\end{equation}


We thus have that, for all $\vx\in \mathcal{X}$:
\begin{align}
\big\| h_\vt(\vx + \delta(\vx)) - h_\vt(\vx) \big\|_2
&\leq
 \big\| J_{h_\vt}(\vx)\,\delta(\vx) \big\|_2  + o(r) \\
&\leq
 \big\| J_{h_\vt}(\vx) \big\|_F \,\big\|\delta(\vx)\big\|_2  + o(r) & &\text{by Cauchy-Schwartz},\\
&\leq r \cdot \big\| J_{h_\vt}(\vx) \big\|_F   + o(r) \label{eq} & &\text{by Eq.~\eqref{eq:bound}}.
\end{align}
Hence,    it follows that:
\begin{align}\label{eq: J fro bound}
\mathbb{E} \left[ \big\| h_\vt(X + \delta(X)) - h_\vt(X) \big\|_2 \right]
&\leq
r \cdot \mathbb{E} \left[ \big\| J_{h_\vt}(X) \big\|_F \right] + o(r)\\
&\leq
\frac{r R B \sqrt{k s}(L R + \|f_\vp(0)\|_2)}{\sigma^2 K_{\mathcal{F}} K_{\mathcal{G}}}
\cdot \operatorname{HSIC}(X, \vz_s) + o(r),
\end{align}
where the last inequality follows from \Cref{eq: HSIC final bound} and  the fact $\frac{\epsilon_R}{\sigma^2} = o(1)$, due to the exponentially decaying term.
\end{proof}

\section{Proof of Lemmas used in Theorem~\ref{thm:hsic-robustness}}
\label{app: lem for thm}
\subsection{Proof of Lemma~\ref{lem:mask-sparsity-bound}} \label{pf:mask-sparsity-bound}
\begin{proof}
Let $\vW_n \in \mathbb{R}^{k \times s}$ and $\vM_s \in \mathbb{R}^{s \times m}$ be the pruned matrices selecting the active coordinates corresponding to the $s$ nonzero entries of the mask. Then the function can equivalently be rewritten as
\begin{equation}
h(\vx) = \vW_s \vM_s f_\vp(\vx).
\end{equation}

Since $f_\vp$ is $L$-Lipschitz and $\|\vx\|_2 \le R$, it follows that
\begin{equation}
\|f_\vp(\vx) - f_\vp(\mathbf{0})\|_2 \le L R,
\end{equation}
thus
\begin{equation}
\|f_\vp(\vx)\|_2 \le \|f_\vp(\mathbf{0})\|_2 + L R, \quad \forall \vx \in \mathcal{X}.
\end{equation}
The masking operation $\vM$ selects $s$ coordinates from $f_\vp(\vx)$, and can be equivalently represented via $\vM_s \in \mathbb{R}^{s \times m}$ as a selector matrix with exactly one nonzero entry per row and no more than one nonzero per column. Then
\begin{equation}
\|\vM_s f_\vp(\vx)\|_2 \le \|f_\vp(\vx)\|_2.
\end{equation}

The corresponding reduced weight matrix $\vW_s \in \mathbb{R}^{k \times s}$ selects the columns of $\vW$ associated with the active coordinates. Since $\|\vW_{ij}\|_\infty \le B$, it follows that
\begin{equation}
\|\vW_s\|_F \le B \sqrt{k s},
\end{equation}
and thus
\begin{equation}
\|\vW_s\|_{2\to2} \le \|\vW_s\|_F \le B \sqrt{k s}.
\end{equation}

where $\lVert \cdot \rVert_F$ is the Frobenius norm. Hence, for any $\vx \in \mathcal{X}$,
\begin{equation}
\|h(\vx)\|_2 = \|\vW_s \vM_s f_\vp(\vx)\|_2 \le \|\vW_s\|_{2\to2} \|\vM_s f_\vp(\vx)\|_2 \le B \sqrt{k s} (L R + \|f_\vp(\mathbf{0})\|_2).
\end{equation}
Taking the supremum over $\vx \in \mathcal{X}$ concludes the proof.
\end{proof}

\subsection{Proof of Corollary~\ref{cor: bounded-nn-ball}} \label{pf: cor bounded nn ball}
\begin{proof}
    As we have shown in \Cref{lem:mask-sparsity-bound}, we have:
    \[\sup_{\vx \in \mathcal{X}} \|h(\vx)\|_2 \le B \sqrt{k s} (L R + \|f_\vp(\mathbf{0})\|_2).\]
    
    Moreover, since the NN can be expressed as $h_\vt(\cdot) = \vW \vM f_\vp(\cdot)$, we have:
    \[\sup_{\vx \in \mathcal{X}} \|\vW \vM f_\vp(\vx)\|_2 \le B \sqrt{k s} (L R + \|f_\vp(\mathbf{0})\|_2).\]

    Therefore, we can upper bound $g_\vw$ as:
    \[\sup_{\vz \in \mathcal{Z}} \|g_\vw(\vz)\|_2= \sup_{\vz \in \mathcal{Z}} \|\vW \vM \vz\|_2 \le B \sqrt{k s} (L R + \|f_\vp(\mathbf{0})\|_2).\]

    Moreover, as $g_\vW$ is a continuous function, we finish the proof.
\end{proof}

\subsection{Proof of Lemma~\ref{lemma:rescaled RKHS equality}}\label{pf: rescaled RKHS equality}
\begin{proof}
\textbf{Step 1.} $\overline{\mF} \subseteq \mathcal{C}_b^{N_\mX}$

Since $k_x$ is continuous and $\mX$ is compact, it follows from Lemma 4.28 of \citet{SVMbook} that all $f \in \mF$ are bounded and continuous. Hence, for any $f \in \mF$, we have
\begin{equation}
    \left\| \frac{N_\mX}{K_\mF} f \right\|_{\infty,2}
    = \frac{N_\mX}{K_\mF} \|f\|_{\infty,2}
    \le N_\mX.
\end{equation}
This implies that every function in $\hat{\mF}$ belongs to $\mathcal{C}_b^{N_\mX}$. Since $\mathcal{C}_b^{N_\mX}$ is closed in the $\|\cdot\|_{\infty,2}$ norm, it follows that
\begin{equation}
   \overline{\mF}\subseteq \mathcal{C}_b^{N_\mX}.
\end{equation}

\textbf{Step 2.} $\mathcal{C}_b^{N_\mX} \subseteq \overline{\mF} $

Let $g \in \mathcal{C}_b^{N_\mX}$. Define $h := \frac{K_\mF}{N_\mX} g$. Then,
\begin{equation}
    \|h\|_{\infty,2} = \frac{K_\mF}{N_\mX} \|g\|_{\infty,2} \le K_\mF,
\end{equation}
so $h \in \mathcal{C}_b(\mX, \mathbb{R}^k)$ and is bounded in sup-norm. Since $\mF$ is universal by Assumption.~\ref{asm: bounded KF}, it is dense in $\mathcal{C}_b(\mX, \mathbb{R}^k)$ under the $\|\cdot\|_{\infty,2}$ norm. Therefore, there exists a sequence $\{f_n\} \subset \mF$ such that
\begin{equation}
    \lim_{ n \to \infty}\|f_n - h\|_{\infty,2} = 0.
\end{equation}
Define the corresponding rescaled sequence $\hat{f}_n := \frac{N_\mX}{K_\mF} f_n \in \hat{\mF}$, and set $\hat{g} := \frac{N_\mX}{K_\mF} h = g$. Computing the limit, 
\begin{equation}
   \lim_{ n \to \infty} \|\hat{f}_n - \hat{g}\|_{\infty,2}
    =\lim_{ n \to \infty} \left\| \frac{N_\mX}{K_\mF} (f_n - h) \right\|_{\infty,2}
    = \lim_{ n \to \infty}\frac{N_\mX}{K_\mF} \|f_n - h\|_{\infty,2} = 0.
\end{equation}
Thus, $g = \hat{g} \in \overline{\mF}$.
\end{proof}

\subsection{Proof of Lemma~\ref{lemma:scaling-sup-cov-sum}}
\label{pf: scaling-sup-cov-sum}
\begin{proof}
We show that 
\[\sum_{j,i} \sup_{f_j, g_i} \operatorname{Cov}[f_j(\vx), g_i(\vz_s)] 
\leq M_{\mF} M_{\mG} 
\sum_{j,i} \sup_{\tilde{f}_j, \tilde{g}_i} \operatorname{Cov}[\tilde{f}_j(\vx), \tilde{g}_i(\vz_s)],\] 

For each pair $(j,i)$, let $\{f_n^{(j)}\} \subset \mF_j$ and $\{g_n^{(i)}\} \subset \mG_i$ be sequences converging to the limit
\begin{equation}
\lim_{n \to \infty} \operatorname{Cov}[f_n^{(j)}(\vx), g_n^{(i)}(\vz_s)] 
= \sup_{f_j, g_i} \operatorname{Cov}[f_j(\vx), g_i(\vz_s)].    
\end{equation}

Define the rescaled sequences:
\begin{equation}
\tilde{f}_n^{(j)} := \frac{1}{M_{\mF}} f_n^{(j)}, \quad 
\tilde{g}_n^{(i)} := \frac{1}{M_{\mG}} g_n^{(i)}.
\end{equation}

Then, by the bilinearity of the covariance operator, we have
\begin{equation}
\operatorname{Cov}[f_n^{(j)}(\vx), g_n^{(i)}(\vz_s)] = M_{\mF} M_{\mG} \operatorname{Cov}[\tilde{f}_n^{(j)}(\vx), \tilde{g}_n^{(i)}(\vz_s)],
\end{equation}
and taking the limit:
\begin{align}
\sup_{f_j, g_i} \operatorname{Cov}[f_j(\vx), g_i(\vz_s)] 
& = \lim_{n \to \infty} \operatorname{Cov}[f_n^{(j)}(\vx), g_n^{(i)}(\vz_s)] \\
& = M_{\mF} M_{\mG} \lim_{n \to \infty} \operatorname{Cov}[\tilde{f}_n^{(j)}(\vx), \tilde{g}_n^{(i)}(\vz_s)]
\le M_{\mF} M_{\mG} \sup_{\tilde{f}_j, \tilde{g}_i} \operatorname{Cov}[\tilde{f}_j(\vx), \tilde{g}_i(\vz_s)].
\end{align}

Summing over all $i, j$ yields the result.

Furthermore, the reverse inequality follows from the same argument.
\end{proof}

\subsection{Proof of Lemma~\ref{lem:bounded-stein-gradient-bound}}
\label{pf: bounded-stein-gradient-bound}
\begin{proof}

\textbf{Step 1 (Integration by Parts).}  
Let $ \phi(\vx) = \exp\left( -\frac{\|\vx\|^2}{2\sigma^2} \right) $, and define $ f(\vx) = h(\vx) \phi(\vx) $.
Applying the product rule:
\begin{equation}
\frac{\partial}{\partial x_i} f(\vx) = \frac{\partial h(\vx)}{\partial x_i} \phi(\vx) + h(\vx) \frac{\partial \phi(\vx)}{\partial x_i}.
\end{equation}
Rearranging:
\begin{equation}
\frac{\partial h(\vx)}{\partial x_i} \phi(\vx) = \frac{\partial}{\partial x_i} (h(\vx)\phi(\vx)) - h(\vx) \frac{\partial \phi(\vx)}{\partial x_i}.
\end{equation}
Integrating over $ \mathcal{X}_R $ and applying the divergence theorem gives:
\begin{equation}
\int_{\mathcal{X}_R} \frac{\partial}{\partial x_i}(h(\vx)\phi(\vx))\,dx = \int_{\partial \mathcal{X}_R} h(\vx) \phi(\vx) \nu_i(\vx)\, dS(\vx),
\end{equation}
where $ \nu(\vx) = \frac{\vx}{\|\vx\|} $ is the outward unit normal and $ \nu_i(\vx) = \frac{x_i}{R} $.

Thus:
\begin{equation}
\int_{\mathcal{X}_R} \frac{\partial h(\vx)}{\partial x_i} \phi(\vx)\,dx
= \int_{\partial \mathcal{X}_R} h(\vx) \phi(\vx) \nu_i(\vx)\, dS(\vx) + \int_{\mathcal{X}_R} h(\vx) \frac{x_i}{\sigma^2} \phi(\vx)\, dx.
\end{equation}

\textbf{Step 2 (Pass to Expectation Form).}  
Dividing through by $ C=\int_{\|x\|\leq R} \phi(\vx) \, dx $, the normalization constant, gives:
\begin{equation}
\mathbb{E} \left[ \frac{\partial h(\vx)}{\partial x_i} \right]
= \frac{1}{\sigma^2} \mathbb{E} \left[ X_i h(\vx) \right]
+ \frac{1}{C} \int_{\|x\|=R} h(\vx) \phi(\vx) \nu_i(\vx)\, dS(\vx).
\end{equation}

\textbf{Step 3 (Bounding the Boundary Term).}  
Since $ |\nu_i(\vx)| \leq 1 $ and $ |h(\vx)| \leq \frac{N_\mX}{\sqrt{k}} $ on $ \mathcal{X}_R $, we have:
\begin{equation}
\left| \int_{\|x\|=R} h(\vx) \phi(\vx) \nu_i(\vx)\, dS(\vx) \right| 
\leq \frac{N_\mX}{\sqrt{k}} \exp\left( -\frac{R^2}{2\sigma^2} \right) \int_{\|x\|=R} dS(\vx).
\end{equation}
The surface area of the sphere is:
\begin{equation}
\int_{\|x\|=R} dS(\vx) = C_d R^{d-1},
\end{equation}
where the constant $ C_d>0$ is the surface area of a d-dimensional unit sphere, depending only on $ d $.

Thus, we can bound the error term as
\begin{equation}
|\epsilon_{i,j}(R)| \leq \frac{N_\mX C_d R^{d-1}}{\sqrt{k}C} \exp\left( -\frac{R^2}{2\sigma^2} \right).
\end{equation}

Then, as $ R \to \infty $, we have:
\begin{equation}
k d \cdot \frac{N_\mX C_d R^{d-1}}{\sqrt{k}C}\exp\left( -\frac{R^2}{2\sigma^2} \right) = o(1),
\end{equation}
since the exponential decay dominates polynomial growth.

\end{proof}

\section{Proof of Lemma~\ref{lem:salient-volume}}
\label{app: pf of lm}
\begin{proof}
Under the same assumptions as in \Cref{thm:hsic-robustness},  recall the HSIC bound on gradients in \cref{eq: HSIC final bound} as 
\begin{equation}
\sum_{j=1}^k \sum_{i=1}^d  \left|\mathbb{E} \left[ \frac{\partial h^{(j)}_\vt(\vx)}{\partial x_i} \right]\right|
\le \frac{1}{\sigma^2} \left( \frac{N_\mX N_{\mathcal{Z}}}{K_{\mathcal{F}} K_{\mathcal{G}}} \cdot \operatorname{HSIC}(\vx, \vz_s) + \epsilon_R \right).
\end{equation}

Thus, we can bound the Frobenius norm of the gradient as
\begin{equation}\label{eq: fro norm hsic}
     \mathbb{E}\| \nabla_\vx h_\vt(\vx) \|_F \leq \sum_{j=1}^k \sum_{i=1}^d  \left|\mathbb{E} \left[ \frac{\partial h^{(j)}_\vt(\vx)}{\partial x_i} \right]\right| \le \frac{1}{\sigma^2} \left( \frac{N_\mX N_{\mathcal{Z}}}{K_{\mathcal{F}} K_{\mathcal{G}}} \cdot \operatorname{HSIC}(\vx, \vz_s) + \epsilon_R \right).
\end{equation}

By Markov inequality, we have
\begin{equation}\label{eq: markov}
    \mathbb{P}(\| \nabla_\vx h_\vt(\vx) \|_F >\epsilon) \leq \frac{1}{\epsilon} \mathbb{E}\left[ \| \nabla_\vx h_\vt(\vx) \|_F  \right]
\end{equation}

Thus, plugging \cref{eq: fro norm hsic} to \cref{eq: markov}, we have
\begin{equation}
    \mathbb{P}(\| \nabla_\vx h_\vt(\vx) \|_F >\epsilon) \leq \frac{1}{\epsilon\sigma^2} \left( \frac{N_\mX N_{\mathcal{Z}}}{K_{\mathcal{F}} K_{\mathcal{G}}} \cdot \operatorname{HSIC}(\vx, \vz_s) + \epsilon_R \right).
\end{equation}

As the error term \(\epsilon_R=O(e^{-\frac{R^2}{2\sigma^2}})\), we have 
\begin{equation}
    \mathbb{P}(\| \nabla_\vx h_\vt(\vx) \|_F >\epsilon) \leq \frac{1}{\epsilon\sigma^2} \left( \frac{N_\mX N_{\mathcal{Z}}}{K_{\mathcal{F}} K_{\mathcal{G}}} \cdot \operatorname{HSIC}(\vx, \vz_s) \right)+ o(1) .
\end{equation}

Substituting the $N_\mX, N_\mZ$, we finish the proof.
\end{proof}

\section{Reproducibility Details}
\label{app:reproducibility}

\subsection{Datasets}
\label{app:datasets}

\begin{table}[t]
\caption{License and source compliance for each dataset.}
\label{tab:dataset_licenses}
\centering
\resizebox{\textwidth}{!}{%
\begin{tabular}{lcccc}
\hline
Dataset & URL & License \\
\hline
ImageNet-1K \citep{imagenet} &
  \href{http://www.image-net.org}{image-net.org} &
  ImageNet Terms \\
ImageNet-9 \citep{imagenet_9} &
  \href{https://github.com/MadryLab/backgrounds_challenge}{GitHub} &
  Inherits ImageNet Terms  \\
COCO \citep{coco} &
  \href{https://cocodataset.org}{cocodataset.org} &
  CC BY 4.0 (annotations) / Flickr TOU (images) \\
CounterAnimal \citep{counteranimal} &
  \href{https://counteranimal.github.io}{counteranimal.github.io} &
  Inherits iNaturalist Terms \\
ISIC-2017 \citep{isic2018skin} & \href{https://challenge.isic-archive.com/data/\#2017}{ISIC Challenge} & CC-0\\
C-MNIST &
  (our codebase) &
  Inherits MNIST Terms (CC BY-SA 3.0) \\
\hline
\end{tabular}
}
\end{table}

COCO is a segmentation dataset consisting of labeled images of various species of animals (See Figure~\ref{fig:attack_types}). For our experiments, we utilize a subset of the dataset composed of images drawn from one of four labels. The four species were carefully selected to ensure the largest possible dataset containing images without overlapping labels. Since we  use the dataset for image classification, each sample should belong to one class and thus include animals from one and only one of the four selected classes. During pre-processing, the dataset is resized to 224x224 pixels. Finally,  segmentation information is used to construct 224x224 masks, where  the 0 entries denote the pixels occupied by the animal (salient object) in the original image. These masks specify the portion of the image shielded from adversarial perturbations. The splits are created from the public training data of COCO by splitting them into train (4509 samples), validation (1127 samples) and test (1411 samples).

C-MNIST is a synthetically constructed variant of the original MNIST dataset \citep{mnist}. To generate it, we first load the standard 28x28 single-channel digit images. Subsequently, each sample is randomly paired with another digit using a fixed seed for reproducibility. The two images are concatenated along the width to form a 56x28 composite, then symmetrically zero-padded to a uniform 64x64 resolution. During training and evaluation, we treat only the left-hand digit as the classification target, ensuring each composite image belongs to exactly one class. We use 80\% of the original train split of MNIST as training data and 20\% as validation data. For testing we use the test set of MNIST, where we also create image pairs as described above.

\begin{figure}[t]
  \centering  \includegraphics[width=0.9\textwidth]{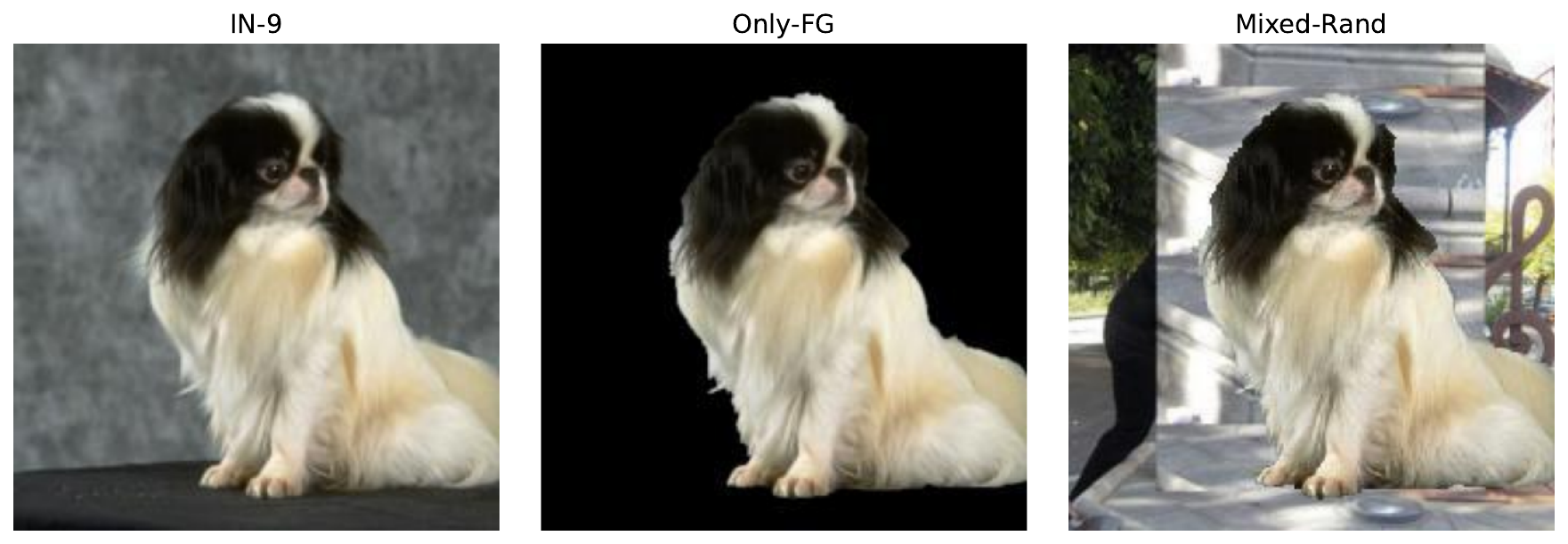}
  \caption{Samples from the three ImageNet9 variations: IN-9 (original), Only-FG, and Mixed-Rand.}
  \label{fig:imagenet9}
\end{figure}

To assess whether \algname{} attends preferentially to salient objects rather than background cues, we evaluate it on two complementary benchmarks: (1) CounterAnimal (CA) \citep{counteranimal}, which splits iNaturalist wildlife photos into a Common set (exhibiting typical backgrounds) and a Counter set (featuring atypical yet plausible backgrounds, see Figure~\ref{fig:counteranimal}), and (2) ImageNet-9 (IN-9) \citep{imagenet_9}, defined as a subset of ImageNet-1K consisting of 368 categories, instantiated in three distinct variants: \emph{Original} images, a \emph{MixedRand} variant in which object foregrounds are transposed onto random-class backgrounds, and an \emph{Only-FG} variant with backgrounds entirely removed (See Figure~\ref{fig:imagenet9}).

Table \ref{tab:dataset_licenses} provides each dataset's source URL and applicable licensing terms.

\subsection{Implementation}
\label{app:training}
The HBaR code was adapted from the original codebase, which is publicly available at \href{https://github.com/neu-spiral/HBaR}{GitHub} (under MIT License). For weight decay, we reuse the PyTorch \citep{Pytorch} implementation and pass it directly to the optimizer. We re-implemented the following regularization methods: Group Lasso Weights, Group Lasso Activations, L1 Sparse Activations, and L1 Sparse Weights. For both the Projected Gradient Descent (PGD) \citep{pgd} and AutoAttack (AA) \citep{autoattack} adversaries, we utilize our version of the TorchAttacks \citep{torchattacks} library, that is adapted for masked attacks.

\subsection{Software and Hardware Setup}
\label{app:hardware}
We built our pipeline in Python, leveraging the PyTorch \citep{Pytorch} library. To conduct our experiments, we use two identical internal servers running Ubuntu 22.04.3 LTS (``Jammy Jellyfish'') on a 5.15.0-84 x86\_64 kernel. Each server is equipped with two Intel Xeon Gold 6326 processors (16 cores each, hyper-threaded for a total of 64 logical CPUs), 512 GiB of RAM, and a single NVIDIA A100 80 GB GPU. For the ablation studies and experiments conducted on the ISIC-2017 dataset, we additionally made use of EuroHPC compute resources, including MareNostrum (BSC, Spain), MeluXina (LuxProvide, Luxembourg), Deucalion (MACC, Portugal), and Discoverer (Sofia Tech, Bulgaria).

\subsection{Hyperparameters}
\label{app:hyperparameters}

We divide our hyperparameters into two groups: those shared by all models, and those tuned or adapted per method and dataset.  
\paragraph{Shared parameters.}
All ImageNet-1K ResNet-50 and COCO ResNet-18 experiments use the Adam optimizer \citep{adam_optimizer} with
\(\beta_1 = 0.9, \beta_2 = 0.999, \epsilon=10^{-8}.\)
We perform an initial grid search on a vanilla ResNet-18, sweeping the learning rate over $\{10^{-3},10^{-4},10^{-5},10^{-6}\}$ in logarithmic steps, and select $\mathrm{LR}=5\times10^{-4}$ for all subsequent runs. The batch size is set to 256, and weight decay is 0 by default (except in weight‐decay experiments). For the C-MNIST experiments we use LeNet-3 \citep{lenet3} with a 1024 embedding space, (as in Figure~\ref{fig:motivation}), we use the learning rate of $\mathrm{LR}=1\times10^{-5}$ and train for 50 epochs from random initialization.

For both COCO and ImageNet-1K we use TorchVision \citep{torchvision} augmentations. Training augmentations include (1) Random Resized Crop to a $224\times224$ patch (scaling and cropping with a random area and aspect ratio), (2) Color Jitter applied with probability $p=0.8$ (brightness $\pm$40\%, contrast $\pm$40\%, saturation $\pm$20\%, hue $\pm$10\%), (3) Random Grayscale with $p=0.2$, (4) Random Horizontal Flip with $p=0.5$, (5) Random Solarize with threshold 0.5 and $p=0.2$, followed by (6) \texttt{ToTensor} and (7) Normalization using per-channel means and standard deviations (ImageNet defaults $[0.485,0.456,0.406]$, $[0.229,0.224,0.225]$ or COCO-computed statistics). At test time, inputs are first resized so that the shorter side is 256 px, then center-cropped to $224\times224$, and finally passed through \texttt{ToTensor} and the same Normalization.

\paragraph{Tuning strategy and per‐method tuning ranges.}
We employ identical hyperparameter tuning strategies for \algname{} and all comparison methods. 30\% of the training corpus is randomly sampled for ImageNet, while the complete training set is used for all other datasets. In either case, 20\% of the samples are used to constitute the validation set. Hyperparameters are optimized via grid search by selecting the model configuration exhibiting the highest robust validation accuracy at the end of training, in which robustness is measured with respect to Projected Gradient Descent (PGD) \citep{pgd} attacks applied to the entire image, so no knowledge of salient or non-salient regions is used. We use dataset-specific perturbation budgets of $\epsilon=\frac{1}{255}$ for ImageNet and $\epsilon=\frac{2}{255}$ for COCO. These values were chosen to be strong enough to select for more robust models, while at the same time being not too strong to induce model collapse to random accuracy, so we can use the metric for model selection. Based on this selection criterion we trained each method on COCO three times and selected the run with highest robust accuracy for validation. For ImageNet we only trained one run. Importantly, no information pertaining to the salient or non-salient regions is leveraged during the tuning. Table~\ref{tab:hyperparams} summarizes the grid ranges we search for each method on ImageNet-1K and COCO.  
\begin{table}[t]
\caption{Hyperparameter tuning ranges per method and dataset.}
\label{tab:hyperparams}
\centering
\begin{tabular}{@{}llccc@{}}
\toprule
\textbf{Method} & \textbf{Parameter} & \textbf{ImageNet-1K} & \textbf{COCO} \\
\midrule
Weight Decay
  & \(\lambda_{\text{wd}}\)
  & \(\{10^{-1},10^{-2},\dots,10^{-6}\}\)
  & \(\{10^{-1},10^{-2},\dots,10^{-6}\}\) \\

L1 Sparse Activations 
  & \(\lambda_{\text{act}}\)
  & \(\{10^{-1},10^{-2},\dots,10^{-6}\}\)
  & \(\{10^{-1},10^{-2},\dots,10^{-6}\}\) \\

Group Lasso Activations
  & \(\lambda_{\text{act}}\)
  & \(\{10^{-1},10^{-2},\dots,10^{-6}\}\)
  & \(\{10^{-1},10^{-2},\dots,10^{-6}\}\) \\

L1 Sparse Weights
  & \(\lambda_{\text{weights}}\)
  & \(\{10^{-1},10^{-2},\dots,10^{-6}\}\)
  & \(\{10^{-1},10^{-2},\dots,10^{-6}\}\) \\

Group Lasso Weights 
  & \(\lambda_{\text{weights}}\)
  & \(\{10^{-1},10^{-2},\dots,10^{-6}\}\)
  & \(\{10^{-1},10^{-2},\dots,10^{-6}\}\) \\

HBaR
  & \(\lambda_x\)
  & \(\{0, 0.001, 0.005, 0.01\}\)
  & \(\{0, 0.001, 0.005, 0.01\}\)\\

\quad
  & \(\lambda_y\)
  & \(\{0, 0.005, 0.05, 0.5\}\)
  & \(\{0, 0.001, 0.01, 0.1\}\) \\

\quad
  & \(\sigma\)
  & \(\{0.5, 1.0, 5.0\}\)
  & \(\{0.5, 5.0\}\) \\

\algname{}
  & \(\lambda_{s}\)
  & \(\{0,0.1,0.5,1.0\}\)
  & \(\{0,0.1,0.5,1.0\}\) \\

\quad 
  & \(\lambda_{n}\)
  & \(\{0,0.05,0.1\}\)
  & \(\{0,0.05,0.1,1.0\}\) \\

\quad 
  & \(\rho_{s}\)
  & \(\{0,0.1,0.15,0.5\}\)
  & \(\{0,0.1,0.15,0.5\}\) \\

\quad 
  & \(\rho_{n}\)
  & \(\{0,0.05,0.1,0.15,0.2,0.3\}\)
  & \(\{0,0.05,0.1,0.15,0.2,0.3\}\) \\

\bottomrule
\end{tabular}
\end{table}

\paragraph{\algname{} selected settings.}
After tuning as above, the final hyperparameters chosen for \algname{} on each dataset are listed in Table~\ref{tab:algname-params}. In all \algname{} runs we also set $\lambda_{ce}=10$ to balance the cross‐entropy scale. Due to the scale of ImageNet-1K, we introduce two scheduling parameters: 
(i) $\beta_{\mathrm{init\_fraction}}$, the fraction of training data used to compute the initial mask values (20\% for ImageNet-1K, 100\% for COCO), and 
(ii) $\beta_{\mathrm{update\_fraction}}$, which determines the amount of training data that must be processed before updating the masks (5\% of the dataset for ImageNet-1K, corresponding to multiple updates per epoch; 100\% for COCO, corresponding to one update per epoch).

\begin{table}[t]
\centering
\caption{\algname{} best hyperparameters for ImageNet-1K and COCO.}
\label{tab:algname-params}
\begin{tabular}{lcc}
\toprule
\textbf{Parameter} & \textbf{ImageNet-1K} & \textbf{COCO} \\
\midrule
$\beta_{\mathrm{init\_fraction}}$    & 20\%    & 100\%   \\
$\beta_{\mathrm{update\_fraction}}$  & 5\%     & 100\%   \\
$\beta_{\mathrm{step}}$              & 0.995   & 0.8     \\
$\lambda_{s}$                        & 0.1     & 0.1     \\
$\lambda_{n}$                        & 0.1     & 0.2    \\
$\rho_{s}$                           & 0.1     & 0.5     \\
$\rho_{n}$                           & 0.1     & 0.05    \\
Shared space variation               & 0.1     & 0.025   \\
\bottomrule
\end{tabular}
\end{table}

\paragraph{Baseline feature selection/regularisation methods.}
Finally, Table~\ref{tab:best-sparsity-params} reports the single best regularization strength found for each of the comparison methods.

\begin{table}[t]
\centering
\caption{Best parameters per comparison method.}
\label{tab:best-sparsity-params}
\begin{tabular}{@{}llcc@{}}
\toprule
\textbf{Method} & \textbf{Parameter} & \textbf{ImageNet-1K} & \textbf{COCO} \\
\midrule
Weight Decay
  & \(\lambda_{\mathrm{wd}}\) 
  & \(10^{-5}\) 
  & \(10^{-3}\) \\

L1 Sparse Weights
  & \(\lambda_{\mathrm{weights}}\) 
  & \(10^{-6}\) 
  & \(10^{-5}\) \\

Group Lasso Weights
  & \(\lambda_{\mathrm{weights}}\) 
  & \(10^{-5}\) 
  & \(10^{-3}\) \\

L1 Sparse Activation
  & \(\lambda_{\mathrm{act}}\) 
  & \(10^{-4}\) 
  & \(10^{-6}\) \\

Group Lasso Activation
  & \(\lambda_{\mathrm{act}}\) 
  & \(10^{-4}\) 
  & \(10^{-5}\) \\

HBaR
  & \(\lambda_x\) 
  & \(10^{-3}\) 
  & \(5^{-3}\) \\

\quad
  & \(\lambda_y\) 
  & \(10^{-1}\) 
  & \(5^{-2}\) \\

\quad
  & \(\sigma\) 
  & \(5\) 
  & \(5\) \\
\bottomrule
\end{tabular}
\end{table}

\section{Additional Experimental Results}
\label{Ap:Time}
\subsection{Training Time Comparison}
\paragraph{COCO Dataset.} The COCO experiments presented in the previous sections were executed for a total of 300 epochs on different machines with varying background workloads. To compare and report the computational intensity of the different training methods, the experiments were repeated on a single machine, albeit for a reduced number of epochs. Table \ref{tab:timing_experiments} provides the average training time per epoch for the various training methods employed on the coco dataset. Those methods were executed for only 20 epochs on the same A100 GPU with 100 GB of memory, whence the average time per epoch was computed. 


\begin{table}[!t]
\caption{Timing experiments on the COCO dataset.}
\label{tab:timing_experiments}
\centering
\begin{tabular}{lcccc}
\hline
Method & average training time per epoch (in seconds)  \\
\hline
Vanilla &  37.121 ($\pm$ 6.032)\\
Weight Decay &  37.687 ($\pm$ 6.680)\\
Group Lasso Weights &  37.353 ($\pm$ 6.151)\\
Group Lasso Activations &  37.354 ($\pm$ 6.455)\\
L1 Sparse Activations & 37.564 ($\pm$ 6.494)\\
L1 Sparse Weights & 37.450 ($\pm$ 6.217)\\
HBaR & 45.739 ($\pm$ 5.895)\\
 \algname{} & 42.852 ($\pm$ 9.672)\\
\hline
\end{tabular}
\end{table}
\begin{table}[!t]
\centering
\caption{Runtime of ImageNet-1K experiments.}
\label{tbl:experiment_runtime}
\begin{tabular}{lc}
\toprule
Method                         & Runtime \\
\midrule
Vanilla                        & 11h 27m \\
Weight Decay                   & 10h 22m \\
Group Lasso Activations        & 10h 20m \\
Group Lasso Weights            & 10h 25m \\
L1 Sparse Activations          & 9h 53m \\
L1 Sparse Weights              & 10h 30m \\
HBaR                           & 12h 04m\\
\algname{}                        & 14h 6m \\
\bottomrule
\end{tabular}
\end{table}

\paragraph{ImageNet-1K Dataset.} The ImageNet-1K experiments were run for 20 epochs on an internal server (see Appendix~\ref{app:hardware}) that was not exclusively reserved for these trials, leading to varying background workloads. Logging was enabled throughout, with \algname{} performing evaluations on the validation set every 5 epochs (during which additional metrics were recorded), while all other methods were evaluated every 10 epochs. These factors may contribute to \algname{}’s longer runtime. As a result, the runtimes reported in Table~\ref{tbl:experiment_runtime} should be taken only as an overview and rough estimate rather than precise timing measurements. Due to the high computational cost of ImageNet-1K experiments and the lack of exclusive server access, precise timing experiments with exclusive access were conducted only on the COCO dataset (See Table~\ref{tab:timing_experiments}).


\subsection{Hyperparameter Sensitivity on COCO-Animals}
\label{app:sensitivity}
We conduct the sensitivity analysis starting from the best-performing configuration on COCO, varying \(\lambda_s,\lambda_n,\rho_s,\rho_n,\beta_{\text{step}}\) while keeping all other settings fixed. 
In Tables~\ref{tab:sens-ls}--\ref{tab:sens-betastep}, we report clean accuracy (\%), PGD robustness at \(\epsilon\in\{1,2,3\}/255\), and the learned salient dimensionality. 
Increasing \(\lambda_n\) or \(\rho_n\) generally improves robustness and compresses the salient subspace up to a regime where excessive regularization degrades performance. 
\(\rho_s\) yields modest robustness gains with gradual salient-space shrinkage, and \(\lambda_s\) exhibits a mild non-monotonic trend around the optimum. For $\beta_{\text{step}}$, very small values (e.g., 0.1) lead to model collapse, while intermediate values (0.3-0.5) reduce clean accuracy despite moderate robustness gains. Larger values (0.8-0.9) maintain high accuracy, with $\beta_{\text{step}}=0.8$ achieving the best results. Overall, the best configuration attains a salient subspace of 14 (out of 512) dimensions with strong robustness.

\begin{table}[t]
\centering
\caption{Sensitivity to \(\lambda_s\).}
\label{tab:sens-ls}
\resizebox{\textwidth}{!}{
\begin{tabular}{lccccc}
\toprule
\(\lambda_s\) & Clean Acc. (\%) & PGD \(\epsilon{=}1/255\) & PGD \(\epsilon{=}2/255\) & PGD \(\epsilon{=}3/255\) & Salient Dim. \\
\midrule
0.01 & 97.73 & 74.12\(\pm\)0.07 & 58.54\(\pm\)0.33 & 37.59\(\pm\)0.38 & 96 \\
0.05 & 98.37 & 74.47\(\pm\)0.19 & 57.90\(\pm\)0.30 & 37.48\(\pm\)0.33 & 101 \\
0.10 & 98.23 & 76.61\(\pm\)0.15 & 60.03\(\pm\)0.38 & 40.61\(\pm\)0.20 & 108 \\
0.20 & 97.73 & 73.52\(\pm\)0.25 & 57.08\(\pm\)0.15 & 37.07\(\pm\)0.34 & 107 \\
0.50 & 98.02 & 74.06\(\pm\)0.25 & 59.96\(\pm\)0.43 & 41.12\(\pm\)0.38 & 107 \\
1.00 & 97.52 & 74.63\(\pm\)0.23 & 60.75\(\pm\)0.32 & 40.26\(\pm\)0.21 & 51 \\
\bottomrule
\end{tabular}}
\end{table}

\begin{table}[t]
\centering
\caption{Sensitivity to \(\lambda_n\).}
\label{tab:sens-lamn}
\resizebox{\textwidth}{!}{
\begin{tabular}{lccccc}
\toprule
\(\lambda_n\) & Clean Acc. (\%) & PGD \(\epsilon{=}1/255\) & PGD \(\epsilon{=}2/255\) & PGD \(\epsilon{=}3/255\) & Salient Dim. \\
\midrule
0.01 & 97.80 & 72.66\(\pm\)0.17 & 55.02\(\pm\)0.29 & 35.83\(\pm\)0.48 & 199 \\
0.05 & 98.23 & 76.61\(\pm\)0.15 & 60.03\(\pm\)0.38 & 40.61\(\pm\)0.20 & 108 \\
0.10 & 98.09 & 76.40\(\pm\)0.09 & 61.06\(\pm\)0.42 & 40.52\(\pm\)0.30 & 29 \\
0.20 & 97.45 & 79.02\(\pm\)0.11 & 68.01\(\pm\)0.33 & 57.07\(\pm\)0.10 & 14 \\
0.50 & 61.30 & 50.40\(\pm\)0.03 & 43.95\(\pm\)0.32 & 35.07\(\pm\)0.15 & 5 \\
1.00 & 32.74 & 32.74\(\pm\)0.00 & 32.74\(\pm\)0.00 & 32.74\(\pm\)0.00 & 0 \\
\bottomrule
\end{tabular}}
\end{table}

\begin{table}[t]
\centering
\caption{Sensitivity to \(\rho_s\).}
\label{tab:sens-rhos}
\resizebox{\textwidth}{!}{
\begin{tabular}{lccccc}
\toprule
\(\rho_s\) & Clean Acc. (\%) & PGD \(\epsilon{=}1/255\) & PGD \(\epsilon{=}2/255\) & PGD \(\epsilon{=}3/255\) & Salient Dim. \\
\midrule
0.01 & 97.73 & 75.18\(\pm\)0.14 & 59.14\(\pm\)0.27 & 38.24\(\pm\)0.33 & 126 \\
0.05 & 98.37 & 73.56\(\pm\)0.13 & 55.01\(\pm\)0.24 & 32.64\(\pm\)0.31 & 120 \\
0.10 & 97.59 & 74.43\(\pm\)0.18 & 58.31\(\pm\)0.12 & 36.98\(\pm\)0.14 & 116 \\
0.20 & 97.87 & 75.76\(\pm\)0.09 & 59.99\(\pm\)0.17 & 40.18\(\pm\)0.20 & 113 \\
0.50 & 98.23 & 76.61\(\pm\)0.15 & 60.03\(\pm\)0.38 & 40.61\(\pm\)0.20 & 108 \\
1.00 & 98.02 & 77.76\(\pm\)0.12 & 61.35\(\pm\)0.60 & 41.87\(\pm\)0.17 & 104 \\
\bottomrule
\end{tabular}}
\end{table}

\begin{table}[t]
\centering
\caption{Sensitivity to \(\rho_n\).}
\label{tab:sens-rhon}
\resizebox{\textwidth}{!}{
\begin{tabular}{lccccc}
\toprule
\(\rho_n\) & Clean Acc. (\%) & PGD \(\epsilon{=}1/255\) & PGD \(\epsilon{=}2/255\) & PGD \(\epsilon{=}3/255\) & Salient Dim. \\
\midrule
0.01 & 98.09 & 74.59\(\pm\)0.17 & 57.07\(\pm\)0.21 & 37.76\(\pm\)0.29 & 124 \\
0.05 & 98.23 & 76.61\(\pm\)0.15 & 60.03\(\pm\)0.38 & 40.61\(\pm\)0.20 & 108 \\
0.10 & 97.52 & 76.48\(\pm\)0.16 & 59.80\(\pm\)0.30 & 39.33\(\pm\)0.16 & 86 \\
0.20 & 97.52 & 75.41\(\pm\)0.11 & 57.76\(\pm\)0.14 & 37.93\(\pm\)0.44 & 57 \\
0.50 & 32.74 & 32.74\(\pm\)0.00 & 32.74\(\pm\)0.00 & 32.74\(\pm\)0.00 & 0 \\
1.00 & 32.74 & 32.74\(\pm\)0.00 & 32.74\(\pm\)0.00 & 32.74\(\pm\)0.00 & 0 \\
\bottomrule
\end{tabular}}
\end{table}

\begin{table}[t]
\centering
\caption{Sensitivity to \(\beta_{\text{step}}\).}
\label{tab:sens-betastep}
\resizebox{\textwidth}{!}{
\begin{tabular}{lccccc}
\toprule
\(\beta_{\text{step}}\) & Clean Acc. (\%) & PGD \(\epsilon{=}1/255\) & PGD \(\epsilon{=}2/255\) & PGD \(\epsilon{=}3/255\) & Salient Dim. \\
\midrule
0.1  & 32.74  & 32.74\(\pm\)0.00 & 32.74\(\pm\)0.00 & 32.74\(\pm\)0.00 & 0 \\
0.3  & 84.55  & 71.23\(\pm\)0.08 & 63.40\(\pm\)0.12 & 54.16\(\pm\)0.12 & 27 \\
0.5  & 85.12 & 71.20\(\pm\)0.16 & 63.23\(\pm\)0.25 & 54.43\(\pm\)0.26 & 30 \\
0.8  & 97.59  & 78.33\(\pm\)0.21 & 69.45\(\pm\)0.29 & 57.92\(\pm\)0.13 & 14 \\
0.9  & 97.66  & 73.98\(\pm\)0.12 & 57.82\(\pm\)0.46 & 36.68\(\pm\)0.35 & 108 \\
0.99 & 98.44 & 74.50\(\pm\)0.15 & 57.35\(\pm\)0.36 & 33.95\(\pm\)0.32 & 490 \\
\bottomrule
\end{tabular}}
\end{table}

\clearpage
\subsection{Effect of Learning Rate}
\label{app:lr}

To verify the effect of learning rate, we vary the learning rate (LR) around the best hyperparameter setting by powers of ten (see Table~\ref{tab:lr}). 
Results indicate a strong influence on both robustness/accuracy and the learned subspace: LR \(=5\times 10^{-4}\) yields the best overall performance with a compact salient subspace (14/512), lower LRs underfit and produce diffuse salient representations, and a higher LR collapses training.

\begin{table}[t]
\centering
\caption{Learning-rate study on COCO-Animals. Salient/Full is measured in a 512-d feature space.}
\label{tab:lr}
\resizebox{\textwidth}{!}{
\begin{tabular}{lcccccccc}
\toprule
Method & Clean Acc. & PGD \(\epsilon{=}1/255\) & PGD \(\epsilon{=}2/255\) & PGD \(\epsilon{=}3/255\) & AA \(\epsilon{=}1/255\) & AA \(\epsilon{=}2/255\) & AA \(\epsilon{=}3/255\) & Salient/Full \\
\midrule
H\text{-}SPLID LR{=}0.005    & 35.23 & 35.23\(\pm\)0.00 & 35.23\(\pm\)0.00 & 35.23\(\pm\)0.00 & 35.23\(\pm\)0.00 & 35.23\(\pm\)0.00 & 35.23\(\pm\)0.00 & 0 / 512 \\
H\text{-}SPLID LR{=}0.0005   & 97.87 & 80.01\(\pm\)0.15 & 69.44\(\pm\)0.16 & 56.95\(\pm\)0.34 & 75.18\(\pm\)0.08 & 58.64\(\pm\)0.17 & 50.40\(\pm\)0.19 & 14 / 512 \\
H\text{-}SPLID LR{=}0.00005  & 96.53 & 64.89\(\pm\)0.11 & 46.78\(\pm\)0.21 & 28.84\(\pm\)0.24 & 56.17\(\pm\)0.10 & 33.98\(\pm\)0.17 & 23.26\(\pm\)0.23 & 33 / 512 \\
H\text{-}SPLID LR{=}0.000005 & 92.49 & 54.30\(\pm\)0.03 & 35.07\(\pm\)0.23 & 17.53\(\pm\)0.20 & 46.82\(\pm\)0.07 & 21.59\(\pm\)0.16 & 11.92\(\pm\)0.11 & 450 / 512 \\
\bottomrule
\end{tabular}}
\end{table}


\clearpage
\section*{NeurIPS Paper Checklist}

\begin{enumerate}

\item {\bf Claims}
    \item[] Question: Do the main claims made in the abstract and introduction accurately reflect the paper's contributions and scope?
    \item[] Answer: \answerYes{} 
    \item[] Justification: We provide empirical evidence for our claim of learning salient features using adversarial attacks against non-salient features. Our theoretical claims are supported with proofs. We show that the output deviation is bounded by HSIC and the salient space dimensionality and also bound the region that could be vulnerable to attacks with HSIC the the salient space dimensionality.
    \item[] Guidelines:
    \begin{itemize}
        \item The answer NA means that the abstract and introduction do not include the claims made in the paper.
        \item The abstract and/or introduction should clearly state the claims made, including the contributions made in the paper and important assumptions and limitations. A No or NA answer to this question will not be perceived well by the reviewers. 
        \item The claims made should match theoretical and experimental results, and reflect how much the results can be expected to generalize to other settings. 
        \item It is fine to include aspirational goals as motivation as long as it is clear that these goals are not attained by the paper. 
    \end{itemize}

\item {\bf Limitations}
    \item[] Question: Does the paper discuss the limitations of the work performed by the authors?
    \item[] Answer: \answerYes{} 
    \item[] Justification: We provide a dedicated paragraph with limitations in the Conclusion \ref{sec:conclustion_future_work}.
    \item[] Guidelines:
    \begin{itemize}
        \item The answer NA means that the paper has no limitation while the answer No means that the paper has limitations, but those are not discussed in the paper. 
        \item The authors are encouraged to create a separate "Limitations" section in their paper.
        \item The paper should point out any strong assumptions and how robust the results are to violations of these assumptions (e.g., independence assumptions, noiseless settings, model well-specification, asymptotic approximations only holding locally). The authors should reflect on how these assumptions might be violated in practice and what the implications would be.
        \item The authors should reflect on the scope of the claims made, e.g., if the approach was only tested on a few datasets or with a few runs. In general, empirical results often depend on implicit assumptions, which should be articulated.
        \item The authors should reflect on the factors that influence the performance of the approach. For example, a facial recognition algorithm may perform poorly when image resolution is low or images are taken in low lighting. Or a speech-to-text system might not be used reliably to provide closed captions for online lectures because it fails to handle technical jargon.
        \item The authors should discuss the computational efficiency of the proposed algorithms and how they scale with dataset size.
        \item If applicable, the authors should discuss possible limitations of their approach to address problems of privacy and fairness.
        \item While the authors might fear that complete honesty about limitations might be used by reviewers as grounds for rejection, a worse outcome might be that reviewers discover limitations that aren't acknowledged in the paper. The authors should use their best judgment and recognize that individual actions in favor of transparency play an important role in developing norms that preserve the integrity of the community. Reviewers will be specifically instructed to not penalize honesty concerning limitations.
    \end{itemize}

\item {\bf Theory assumptions and proofs}
    \item[] Question: For each theoretical result, does the paper provide the full set of assumptions and a complete (and correct) proof?
    \item[] Answer: \answerYes{} 
    \item[] Justification: We present the assumptions in \Cref{subsec: theory} and the proofs appears in \Cref{app: pf of thm}, \Cref{app: lem for thm}, and \Cref{app: pf of lm}.
    \item[] Guidelines:
    \begin{itemize}
        \item The answer NA means that the paper does not include theoretical results. 
        \item All the theorems, formulas, and proofs in the paper should be numbered and cross-referenced.
        \item All assumptions should be clearly stated or referenced in the statement of any theorems.
        \item The proofs can either appear in the main paper or the supplemental material, but if they appear in the supplemental material, the authors are encouraged to provide a short proof sketch to provide intuition. 
        \item Inversely, any informal proof provided in the core of the paper should be complemented by formal proofs provided in appendix or supplemental material.
        \item Theorems and Lemmas that the proof relies upon should be properly referenced. 
    \end{itemize}

    \item {\bf Experimental result reproducibility}
    \item[] Question: Does the paper fully disclose all the information needed to reproduce the main experimental results of the paper to the extent that it affects the main claims and/or conclusions of the paper (regardless of whether the code and data are provided or not)?
    \item[] Answer: \answerYes{} 
    \item[] Justification: We detail our experimental setup in Section \ref{sec:experiments}. Appendix \ref{app:reproducibility} describes the implementation details, our hyperparameter-tuning procedure, and the exact settings used to train our models. In Table \ref{tab:dataset_licenses}, we provide links to download the datasets used. We are submitting our code alongside the supplementary materials and will open-source it, together with the best-performing models, upon acceptance.

    \item[] Guidelines:
    \begin{itemize}
        \item The answer NA means that the paper does not include experiments.
        \item If the paper includes experiments, a No answer to this question will not be perceived well by the reviewers: Making the paper reproducible is important, regardless of whether the code and data are provided or not.
        \item If the contribution is a dataset and/or model, the authors should describe the steps taken to make their results reproducible or verifiable. 
        \item Depending on the contribution, reproducibility can be accomplished in various ways. For example, if the contribution is a novel architecture, describing the architecture fully might suffice, or if the contribution is a specific model and empirical evaluation, it may be necessary to either make it possible for others to replicate the model with the same dataset, or provide access to the model. In general. releasing code and data is often one good way to accomplish this, but reproducibility can also be provided via detailed instructions for how to replicate the results, access to a hosted model (e.g., in the case of a large language model), releasing of a model checkpoint, or other means that are appropriate to the research performed.
        \item While NeurIPS does not require releasing code, the conference does require all submissions to provide some reasonable avenue for reproducibility, which may depend on the nature of the contribution. For example
        \begin{enumerate}
            \item If the contribution is primarily a new algorithm, the paper should make it clear how to reproduce that algorithm.
            \item If the contribution is primarily a new model architecture, the paper should describe the architecture clearly and fully.
            \item If the contribution is a new model (e.g., a large language model), then there should either be a way to access this model for reproducing the results or a way to reproduce the model (e.g., with an open-source dataset or instructions for how to construct the dataset).
            \item We recognize that reproducibility may be tricky in some cases, in which case authors are welcome to describe the particular way they provide for reproducibility. In the case of closed-source models, it may be that access to the model is limited in some way (e.g., to registered users), but it should be possible for other researchers to have some path to reproducing or verifying the results.
        \end{enumerate}
    \end{itemize}

\item {\bf Open access to data and code}
    \item[] Question: Does the paper provide open access to the data and code, with sufficient instructions to faithfully reproduce the main experimental results, as described in supplemental material?
    \item[] Answer: \answerYes{} 
    \item[] Justification: In Table \ref{tab:dataset_licenses}, we provide links to download the datasets used. The C-MNIST dataset can be constructed using our code. We are submitting our code alongside the supplementary materials and will open-source it, together with the best-performing models, upon acceptance. We include a ``README'' file that details the environment setup, folder and script structure, and execution instructions for our method and all comparison methods.
    
    \item[] Guidelines:
    \begin{itemize}
        \item The answer NA means that paper does not include experiments requiring code.
        \item Please see the NeurIPS code and data submission guidelines (\url{https://nips.cc/public/guides/CodeSubmissionPolicy}) for more details.
        \item While we encourage the release of code and data, we understand that this might not be possible, so “No” is an acceptable answer. Papers cannot be rejected simply for not including code, unless this is central to the contribution (e.g., for a new open-source benchmark).
        \item The instructions should contain the exact command and environment needed to run to reproduce the results. See the NeurIPS code and data submission guidelines (\url{https://nips.cc/public/guides/CodeSubmissionPolicy}) for more details.
        \item The authors should provide instructions on data access and preparation, including how to access the raw data, preprocessed data, intermediate data, and generated data, etc.
        \item The authors should provide scripts to reproduce all experimental results for the new proposed method and baselines. If only a subset of experiments are reproducible, they should state which ones are omitted from the script and why.
        \item At submission time, to preserve anonymity, the authors should release anonymized versions (if applicable).
        \item Providing as much information as possible in supplemental material (appended to the paper) is recommended, but including URLs to data and code is permitted.
    \end{itemize}

\item {\bf Experimental setting/details}
    \item[] Question: Does the paper specify all the training and test details (e.g., data splits, hyperparameters, how they were chosen, type of optimizer, etc.) necessary to understand the results?
    \item[] Answer: \answerYes{} 
    \item[] Justification: We detail our experimental setup in Section \ref{sec:experiments}. Appendix \ref{app:reproducibility} describes implementation details, our hyperparameter-tuning procedure and lists the exact settings used to train our models. Data splits are described in Section \ref{app:datasets} and in Section \ref{subsec:experiment_setup}.
    \item[] Guidelines:
    \begin{itemize}
        \item The answer NA means that the paper does not include experiments.
        \item The experimental setting should be presented in the core of the paper to a level of detail that is necessary to appreciate the results and make sense of them.
        \item The full details can be provided either with the code, in appendix, or as supplemental material.
    \end{itemize}

\item {\bf Experiment statistical significance}
    \item[] Question: Does the paper report error bars suitably and correctly defined or other appropriate information about the statistical significance of the experiments?
    \item[] Answer: \answerYes{} 
    \item[] Justification: We provide standard deviations over attacks for COCO, where our method, \algname{} outperforms other methods by more than two standard deviations (see \Cref{sec:experiments} and \Cref{tab:adversarial_robustness}), but with no dedicated statistical significance testing. The experiments on ImageNet were performed using a single seed due to their computational cost. We used train, val and test splits for all experiments and provide details on data splits in Section \ref{app:datasets}.
    \item[] Guidelines:
    \begin{itemize}
        \item The answer NA means that the paper does not include experiments.
        \item The authors should answer "Yes" if the results are accompanied by error bars, confidence intervals, or statistical significance tests, at least for the experiments that support the main claims of the paper.
        \item The factors of variability that the error bars are capturing should be clearly stated (for example, train/test split, initialization, random drawing of some parameter, or overall run with given experimental conditions).
        \item The method for calculating the error bars should be explained (closed form formula, call to a library function, bootstrap, etc.)
        \item The assumptions made should be given (e.g., Normally distributed errors).
        \item It should be clear whether the error bar is the standard deviation or the standard error of the mean.
        \item It is OK to report 1-sigma error bars, but one should state it. The authors should preferably report a 2-sigma error bar than state that they have a 96\% CI, if the hypothesis of Normality of errors is not verified.
        \item For asymmetric distributions, the authors should be careful not to show in tables or figures symmetric error bars that would yield results that are out of range (e.g. negative error rates).
        \item If error bars are reported in tables or plots, The authors should explain in the text how they were calculated and reference the corresponding figures or tables in the text.
    \end{itemize}

\item {\bf Experiments compute resources}
    \item[] Question: For each experiment, does the paper provide sufficient information on the computer resources (type of compute workers, memory, time of execution) needed to reproduce the experiments?
    \item[] Answer: \answerYes{}
    \item[] Justification: The timing experiments (Appendix \ref{Ap:Time}) provide the relevant information on compute resources. Details on the hardware used are given in Appendix \ref{app:hardware}.
    
    \item[] Guidelines:
    \begin{itemize}
        \item The answer NA means that the paper does not include experiments.
        \item The paper should indicate the type of compute workers CPU or GPU, internal cluster, or cloud provider, including relevant memory and storage.
        \item The paper should provide the amount of compute required for each of the individual experimental runs as well as estimate the total compute. 
        \item The paper should disclose whether the full research project required more compute than the experiments reported in the paper (e.g., preliminary or failed experiments that didn't make it into the paper). 
    \end{itemize}
    
\item {\bf Code of ethics}
    \item[] Question: Does the research conducted in the paper conform, in every respect, with the NeurIPS Code of Ethics \url{https://neurips.cc/public/EthicsGuidelines}?
    \item[] Answer: \answerYes{} 
    \item[] Justification: We read and followed the Code of Ethics.
    \item[] Guidelines:
    \begin{itemize}
        \item The answer NA means that the authors have not reviewed the NeurIPS Code of Ethics.
        \item If the authors answer No, they should explain the special circumstances that require a deviation from the Code of Ethics.
        \item The authors should make sure to preserve anonymity (e.g., if there is a special consideration due to laws or regulations in their jurisdiction).
    \end{itemize}

\item {\bf Broader impacts}
    \item[] Question: Does the paper discuss both potential positive societal impacts and negative societal impacts of the work performed?
    \item[] Answer: 
    \answerYes{} 
    \item[] Justification: We discuss the potential scientific impact of producing salient latent representations in the introduction and related work. We do not see a direct relation between our work with any other societal impacts.
    \item[] Guidelines: 
    \begin{itemize}
        \item The answer NA means that there is no societal impact of the work performed.
        \item If the authors answer NA or No, they should explain why their work has no societal impact or why the paper does not address societal impact.
        \item Examples of negative societal impacts include potential malicious or unintended uses (e.g., disinformation, generating fake profiles, surveillance), fairness considerations (e.g., deployment of technologies that could make decisions that unfairly impact specific groups), privacy considerations, and security considerations.
        \item The conference expects that many papers will be foundational research and not tied to particular applications, let alone deployments. However, if there is a direct path to any negative applications, the authors should point it out. For example, it is legitimate to point out that an improvement in the quality of generative models could be used to generate deepfakes for disinformation. On the other hand, it is not needed to point out that a generic algorithm for optimizing neural networks could enable people to train models that generate Deepfakes faster.
        \item The authors should consider possible harms that could arise when the technology is being used as intended and functioning correctly, harms that could arise when the technology is being used as intended but gives incorrect results, and harms following from (intentional or unintentional) misuse of the technology.
        \item If there are negative societal impacts, the authors could also discuss possible mitigation strategies (e.g., gated release of models, providing defenses in addition to attacks, mechanisms for monitoring misuse, mechanisms to monitor how a system learns from feedback over time, improving the efficiency and accessibility of ML).
    \end{itemize}
    
\item {\bf Safeguards}
    \item[] Question: Does the paper describe safeguards that have been put in place for responsible release of data or models that have a high risk for misuse (e.g., pretrained language models, image generators, or scraped datasets)?
    \item[] Answer: \answerNA{} 
    \item[] Justification:  Our work does not have high-risk of misuse and we do not release data.
    \item[] Guidelines:
    \begin{itemize}
        \item The answer NA means that the paper poses no such risks.
        \item Released models that have a high risk for misuse or dual-use should be released with necessary safeguards to allow for controlled use of the model, for example by requiring that users adhere to usage guidelines or restrictions to access the model or implementing safety filters. 
        \item Datasets that have been scraped from the Internet could pose safety risks. The authors should describe how they avoided releasing unsafe images.
        \item We recognize that providing effective safeguards is challenging, and many papers do not require this, but we encourage authors to take this into account and make a best faith effort.
    \end{itemize}

\item {\bf Licenses for existing assets}
    \item[] Question: Are the creators or original owners of assets (e.g., code, data, models), used in the paper, properly credited and are the license and terms of use explicitly mentioned and properly respected?
    \item[] Answer: \answerYes{} 
    \item[] Justification: We provide licenses and citations of used datasets in Table \ref{tab:dataset_licenses}. We cite original code packages and repositories in Appendix~\ref{app:training}.  
    \item[] Guidelines:
    \begin{itemize}
        \item The answer NA means that the paper does not use existing assets.
        \item The authors should cite the original paper that produced the code package or dataset.
        \item The authors should state which version of the asset is used and, if possible, include a URL.
        \item The name of the license (e.g., CC-BY 4.0) should be included for each asset.
        \item For scraped data from a particular source (e.g., website), the copyright and terms of service of that source should be provided.
        \item If assets are released, the license, copyright information, and terms of use in the package should be provided. For popular datasets, \url{paperswithcode.com/datasets} has curated licenses for some datasets. Their licensing guide can help determine the license of a dataset.
        \item For existing datasets that are re-packaged, both the original license and the license of the derived asset (if it has changed) should be provided.
        \item If this information is not available online, the authors are encouraged to reach out to the asset's creators.
    \end{itemize}

\item {\bf New assets}
    \item[] Question: Are new assets introduced in the paper well documented and is the documentation provided alongside the assets?
    \item[] Answer: \answerYes{} 
    \item[] Justification: We detail model training in Appendix \ref{app:reproducibility} and Section \ref{sec:experiments}. Appendix \ref{app:datasets} describes how the C-MNIST dataset was constructed. Licenses for all assets are specified in Table \ref{tab:dataset_licenses} and \ref{app:training}. The submission and all supplementary materials have been anonymized.
    
    \item[] Guidelines:
    \begin{itemize}
        \item The answer NA means that the paper does not release new assets.
        \item Researchers should communicate the details of the dataset/code/model as part of their submissions via structured templates. This includes details about training, license, limitations, etc. 
        \item The paper should discuss whether and how consent was obtained from people whose asset is used.
        \item At submission time, remember to anonymize your assets (if applicable). You can either create an anonymized URL or include an anonymized zip file.
    \end{itemize}

\item {\bf Crowdsourcing and research with human subjects}
    \item[] Question: For crowdsourcing experiments and research with human subjects, does the paper include the full text of instructions given to participants and screenshots, if applicable, as well as details about compensation (if any)? 
    \item[] Answer: \answerNA{} 
    \item[] Justification: No use of human subjects or crowdsourcing.
    \item[] Guidelines:
    \begin{itemize}
        \item The answer NA means that the paper does not involve crowdsourcing nor research with human subjects.
        \item Including this information in the supplemental material is fine, but if the main contribution of the paper involves human subjects, then as much detail as possible should be included in the main paper. 
        \item According to the NeurIPS Code of Ethics, workers involved in data collection, curation, or other labor should be paid at least the minimum wage in the country of the data collector. 
    \end{itemize}

\item {\bf Institutional review board (IRB) approvals or equivalent for research with human subjects}
    \item[] Question: Does the paper describe potential risks incurred by study participants, whether such risks were disclosed to the subjects, and whether Institutional Review Board (IRB) approvals (or an equivalent approval/review based on the requirements of your country or institution) were obtained?
    \item[] Answer: \answerNA{} 
    \item[] Justification: No use of human subjects or crowdsourcing.
    \item[] Guidelines:
    \begin{itemize}
        \item The answer NA means that the paper does not involve crowdsourcing nor research with human subjects.
        \item Depending on the country in which research is conducted, IRB approval (or equivalent) may be required for any human subjects research. If you obtained IRB approval, you should clearly state this in the paper. 
        \item We recognize that the procedures for this may vary significantly between institutions and locations, and we expect authors to adhere to the NeurIPS Code of Ethics and the guidelines for their institution. 
        \item For initial submissions, do not include any information that would break anonymity (if applicable), such as the institution conducting the review.
    \end{itemize}

\item {\bf Declaration of LLM usage}
    \item[] Question: Does the paper describe the usage of LLMs if it is an important, original, or non-standard component of the core methods in this research? Note that if the LLM is used only for writing, editing, or formatting purposes and does not impact the core methodology, scientific rigorousness, or originality of the research, declaration is not required.
    \item[] Answer: \answerNA{} 
    \item[] Justification: LLMs were only used for writing assistance.
    \item[] Guidelines:
    \begin{itemize}
        \item The answer NA means that the core method development in this research does not involve LLMs as any important, original, or non-standard components.
        \item Please refer to our LLM policy (\url{https://neurips.cc/Conferences/2025/LLM}) for what should or should not be described.
    \end{itemize}

\end{enumerate}

\end{document}